\documentclass[runningheads]{llncs}

\usepackage{eccv}

\usepackage{graphicx}
\usepackage{booktabs}
\usepackage[accsupp]{axessibility}  

\usepackage{makecell}
\usepackage{multirow}
\usepackage{colortbl}
\usepackage{amsmath,pifont}
\usepackage{mathrsfs}
\usepackage{tabularx}
\usepackage{cuted}
\usepackage{comment}

\usepackage{hyperref}

\usepackage{orcidlink}

\definecolor{emp_1}{HTML}{cc2936}
\definecolor{emp_2}{HTML}{157f1f}
\definecolor{emp_3}{HTML}{072ac8}
\definecolor{emp_4}{HTML}{f3bc6a}

\begin{document}

\def\eg{\emph{e.g.},}
\def\etal{\emph{et al.}}
\def\ie{\emph{i.e.},}
\def\etc{\emph{etc.}}
\def\wrt{w.r.t.}
\def\vs{\emph{vs.}}
\def\blue#1{\textcolor{blue}{#1}}
\def\red#1{\textcolor{red}{#1}}
\def\green#1{\textcolor{green}{#1}}

\setlength{\textfloatsep}{13pt plus 5pt minus 5pt}
\setlength{\floatsep}{8pt}
\setlength{\abovecaptionskip}{4pt}
\setlength{\belowcaptionskip}{4pt}

\title{Efficient RGB-T Object Detection via Sparse Cross-Modality Fusion} 


\author{Chao Tian\inst{1}\orcidlink{0000-0002-7029-5065} \and
Zikun Zhou\inst{1}\thanks{Corresponding authors.}\orcidlink{0000-0002-2687-7762} \and
Chao Yang\inst{1}\orcidlink{0009-0000-0426-3826} \and 
Guoqing Zhu\inst{1}\orcidlink{0009-0001-8153-2636} \and \\ 
Zhenyu He\inst{1}\unskip$^{\star}$\orcidlink{0000-0002-2546-8721}
}

\authorrunning{C.~Tian et al.}

\institute{Harbin Institute of Technology, Shenzhen, Guangdong, China \\
\email{tianchao@stu.hit.edu.cn, zhouzikunhit@gmail.com, \{20b951014,20B951002\}@stu.hit.edu.cn, zhenyuhe@hit.edu.cn}
}

\maketitle
\begin{abstract}

RGB-T detectors leverage the complementary strengths of visible and thermal infrared modalities, achieving robust performance under challenging conditions. Many of them resort to heavy dual backbones and exhaustive cross-modality fusion across the entire image, leading to impractically high computational costs. We observe that most image regions are smooth backgrounds (\eg~sky, ground) that can be easily handled by lightweight single-modality models. In light of this observation, we propose a sparse fusion mechanism for efficient RGB-T detection: first rapidly scanning the image to identify the proposals and then carefully examining the remaining sparse proposals via feature fusion. We propose a two-stage framework to instantiate this mechanism, which performs detection in two stages: 1) a lightweight and modality-specific detection stage that produces high-recall RoIs, and 2) a fusion-driven examination and refinement stage that filters out the false positives and refines the bounding boxes. This design enables the detector to adaptively allocate more computational resources to the potential foregrounds, improving the efficiency while ensuring detection accuracy. Extensive experiments show that our method achieves competitive performance with substantially fewer parameters and lower cost, while maintaining strong scalability to high-resolution images.


\end{abstract}

\section{Introduction}
\label{sec:intro}
RGB-Thermal infrared (RGB-T) object detection aims to leverage the complementarity from different modalities to handle various challenging conditions~\cite{liu2016multispectral,gaff}, drawing much attention in recent years. Numerous benchmarks~\cite{zhang2020multispectral_flir,m3fd,llvip} and detectors~\cite{gaff, li2022confidence, Tian_Yang_Zhu_Wang_He_2025} have been proposed. Detectors based on feature fusion have acquired superior detection performance~\cite{shen2024icafusion}, which typically extract the RGB and thermal infrared (TIR) features using two independent backbones and perform pixel-to-pixel feature fusion~\cite{Yang_Liang_Li_Zhang_2025, tian2024cross}. 
However, the computational complexity of these powerful RGB-T detectors remains undesirably high, reaching up to hundreds of GFLOPs~\cite{cft, shen2024icafusion, lee2024crossformer, FusionMamba}, restricting their practical applicability. 

\begin{figure}[!t]
    \centering
    \includegraphics[width=\linewidth]{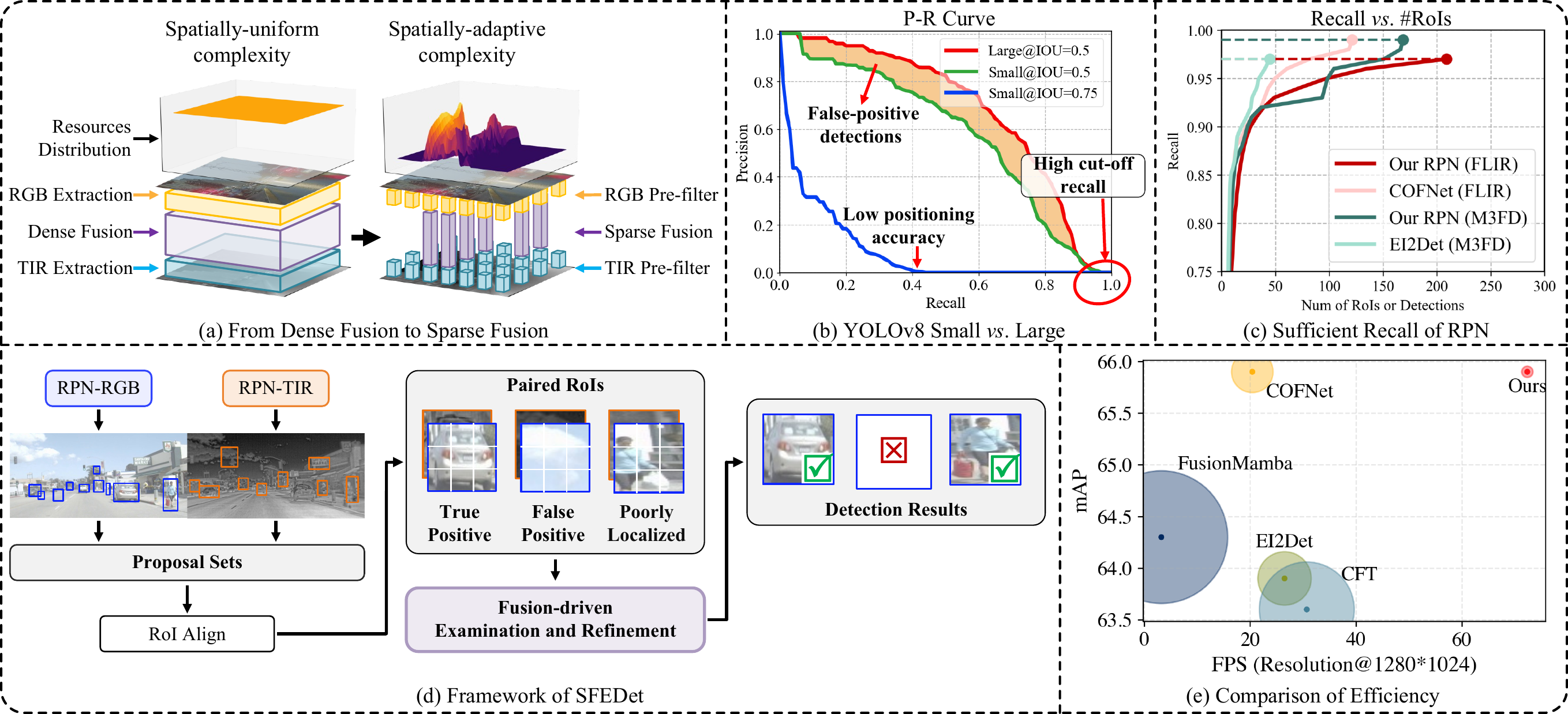}
    \caption{
    Illustration for our motivation and insight. (a) Most of an image is background that can be efficiently filtered out without fusion. Sparse fusion for potential areas could avoid the computationally intensive dense fusion over the entire image.
    (b) The P-R curves of YOLOv8-Large and -Small suggest that, though the lightweight detector has more false positives and lower location accuracy, it maintains a high recall rate for proposals. (c) Our RPNs demonstrate a sufficient recall comparable to that of large detectors, ensuring the feasibility of first extracting proposals and then accordingly implementing sparse fusion.
    (d) SFEDet detects objects by first rapidly scanning the image to identify sparse proposals, followed by a thorough examination and refinement driven by feature fusion. (e) Comparison between our SFEDet and recent RGB-T detectors. The radius of circles indicates the number of parameters. } 
    \label{fig:moti}
\end{figure}

Existing powerful RGB-T detectors primarily employ heavy dual backbones and perform sophisticated feature fusion uniformly across the entire image to detect the target objects accurately. Nevertheless, the backgrounds, such as the sky and ground, which generally occupy the majority of the image, are smooth and can be readily distinguished by lightweight models, even when utilizing only a single modality. Hence, processing the backgrounds with a heavy feature extractor and a complex fusion mechanism is unnecessary and highly inefficient. In light of this, an effective strategy for constructing lightweight yet high-precision RGB-T detectors is to pre-filter these simple background regions using compact single-modality models and employ a powerful cross-modality fusion module to process the remaining regions. This is essentially an efficient sparse fusion mechanism, adaptively allocating more computational resources to potential foregrounds, as shown in Figure~\ref{fig:moti} (a).

The prerequisite for the sparse fusion mechanism is that the lightweight model can identify potential foregrounds at a high recall rate without unacceptable mistaken drops. To figure this out, we conduct an analytical experiment with YOLOv8-Large and YOLOv8-Small~\cite{yolov8}. Figure~\ref{fig:moti} (b) presents the Precision–Recall (PR) curves on FLIR~\cite{zhang2020multispectral_flir}. We observe that the lightweight detector achieves a cut-off recall comparable to that of the heavyweight detector, both remarkably close to 100\%, when the IoU threshold is 50\%. Such a high recall validates the effectiveness of a lightweight model for searching potential foregrounds, \ie~proposals. This demonstrates the feasibility of the sparse fusion mechanism. Nevertheless, we observe two limitations from the PR curves. First, the PR curve of the lightweight detector remains consistently lower than that of the heavyweight one at IoU=50\%. This suggests a high false-positive rate in the preliminary proposals, which therefore requires further verification for accurate recognition. Second, the recall drops sharply at IoU=75\%, indicating the limited localization accuracy of the lightweight model. These two issues can be effectively mitigated with the robust features derived from cross-modality fusion. 

In this paper, we instantiate \textbf{S}parse \textbf{F}usion mechanism and propose an \textbf{E}fficient RGB-T object \textbf{Det}ector, dubbed \textbf{SFEDet}, as illustrated in Figure~\ref{fig:moti}(d). Specifically, we employ two lightweight detectors, one for each modality, serving as region proposal networks (RPNs) to identify potential foreground regions, \ie~Regions of Interest (RoIs), for further detailed examination. 
Herein, to avoid missing real objects during proposal extraction, we construct the proposal set by taking the union of RoIs from individual modalities. Subsequently, we borrow RoI Align~\cite{maskrcnn} to extract paired RoI features from the two modalities. Despite the high recall of the collected proposals extracted by our RPNs, as shown in Figure~\ref{fig:moti} (c), the RPNs still suffer from false positives and inaccurate localization. To accurately filter out false positives and refine coarse bounding boxes, we introduce a \textbf{F}usion-driven \textbf{E}xamination and \textbf{R}efinement (\textbf{FER}) module to process the RoIs for final detection, which performs multi-step feature fusion followed by step-wise classification and box refinement. In this way, the lightweight RPNs process the entire image preliminarily, while the sophisticated FER module operates only on sparse RoI features with fixed spatial sizes. This design enables a favorable balance between efficiency and performance. Moreover, due to the sparse fusion mechanism, SFEDet demonstrates excellent scalability for high-resolution images, as the computational complexity of FER depends solely on the number of RoIs, as demonstrated in Figure~\ref{fig:moti} (e).

The implementation of the FER module is crucial to the efficiency and performance of our SFEDet. To this end, we introduce a Rolling Convolution operator into the FER module to replace the conventional convolution operator, improving its efficiency. Furthermore, we adopt a denoising training strategy to facilitate the FER module to learn to process proposals with large localization errors, consequently improving its detection performance. We conduct elaborate experiments on FLIR, M3FD, and LLVIP datasets, where the SFEDet achieves a balance between efficiency and precision. 
The main contributions can be summarized as follows: 

\begin{itemize}
\item We introduce a sparse fusion mechanism that adaptively allocates greater computational resources to potential foreground regions; it is then instantiated within our SFEDet framework, enabling SFEDet to achieve a favorable balance between efficiency and performance.

\item We propose a fusion-driven examination and refinement method to process the sparse proposals, which effectively filters out the false positives and refines the coarse bounding boxes via step-wise feature enhancement.

\item We conduct extensive experiments on popular RGB-T detection benchmarks, demonstrating the effectiveness of our method and highlighting its scalability for high-resolution images.
\end{itemize}

\section{Related works} 
\subsection{RGB-T Object Detection} 

In recent years, several benchmarks~\cite{kaist,zhang2020multispectral_flir} have been established and promote the development of the detection area. Some of them, such as M3FD~\cite{m3fd} and LLVIP~\cite{llvip}, are well aligned and adopted not only in object detection but also in pixel-level image fusion. 

The algorithms for RGB-T object detection have advanced substantially, evolving from image fusion-based methods~\cite{Wagner_Fischer_Herman_Behnke_2016, Rethink_TIV} and box fusion-based methods~\cite{chen2022multimodal} to feature fusion-based approaches. Among these, He~\etal~\cite{He_Tang_Zou_Zhang_2023} proposed weighting different modalities according to the information difference measured within their local neighborhoods. Li~\etal~\cite{li2022confidence, Kim_uncertainty} adjusted the fusion weights of RGB and TIR features based on illumination conditions. Hu~\etal~\cite{Hu_Edge_2025} incorporated edge priors to constrain the fusion process, while ICAFusion\cite{shen2024icafusion} employed a cross-attention mechanism for modality guidance. Other studies~\cite{gaff, Zhou_Li_Yang_Wei_Pu_Luo_Jia_2025} used box-aligned masks as spatial guidance to achieve more precise fusion.
Chen~\etal~\cite{Chen_igt_2023, Yang_Liang_2025} introduced vision transformers into RGB-T object detectors, while Dong~\cite{Dong_fusion_mamba_2025} employed Mamba~\cite{gu2024mamba} to perform feature fusion.
In addition, several studies have addressed specific challenges in RGB-T object detection. Tian~\etal~\cite{tian2024cross, zhang2019weakly} focused on the non-alignment issue, whereas Zhou~\etal~\cite{zhou2020improving, Tian_Yang_Zhu_Wang_He_2025} investigated the modality imbalance problem, where the RGB and TIR modalities contribute unequally to the fusion results.
However, the computational overhead of feature fusion-based RGB-T detectors has received little attention to date.

\subsection{Efficient Object Detectors} 

Efficient lightweight detectors have been extensively studied in general object detection. The YOLO~\cite{yolo} series is well known for enabling real-time detection. Early versions~\cite{yolov3, bochkovskiy2020yolov4} are anchor-based, which makes both training and deployment relatively complex. Subsequent improvements, such as YOLOX~\cite{yolox} and TOOD~\cite{feng2021tood}, introduce dynamic sample assignment, a technique that has been widely adopted in later versions of YOLO~\cite{yolov8, yolov10}. YOLOv7~\cite{yolov7} and PP-YOLOE~\cite{ppyoloe} further enhance the network architectures to improve inference speed and overall efficiency. YOLOv10~\cite{yolov10} focuses on eliminating the non-maximum suppression (NMS) post-processing step, which has long been considered a bottleneck in detector acceleration.
DETR-like detectors~\cite{zhudeformable,dndetr, dino} generally achieve higher upper-bound performance but at the cost of increased computational complexity. RT-DETR~\cite{RTdetr} is a real-time DETR-based detector that strikes a favorable balance between speed and accuracy.

Although efficient single-modality detectors have been extensively explored, the modal-specific and unshared feature extraction pipelines, together with the subsequent fusion process, remain computationally demanding. This issue has received limited attention in prior studies. Deevi~\cite{deevi_rgbx} attempts to alleviate the training cost by freezing two backbones; however, the overall computational complexity of the model remains unchanged. Addressing this limitation constitutes the primary goal of our work.

\section{Method}

\begin{figure}[!t]
    \includegraphics[width=0.988\linewidth]{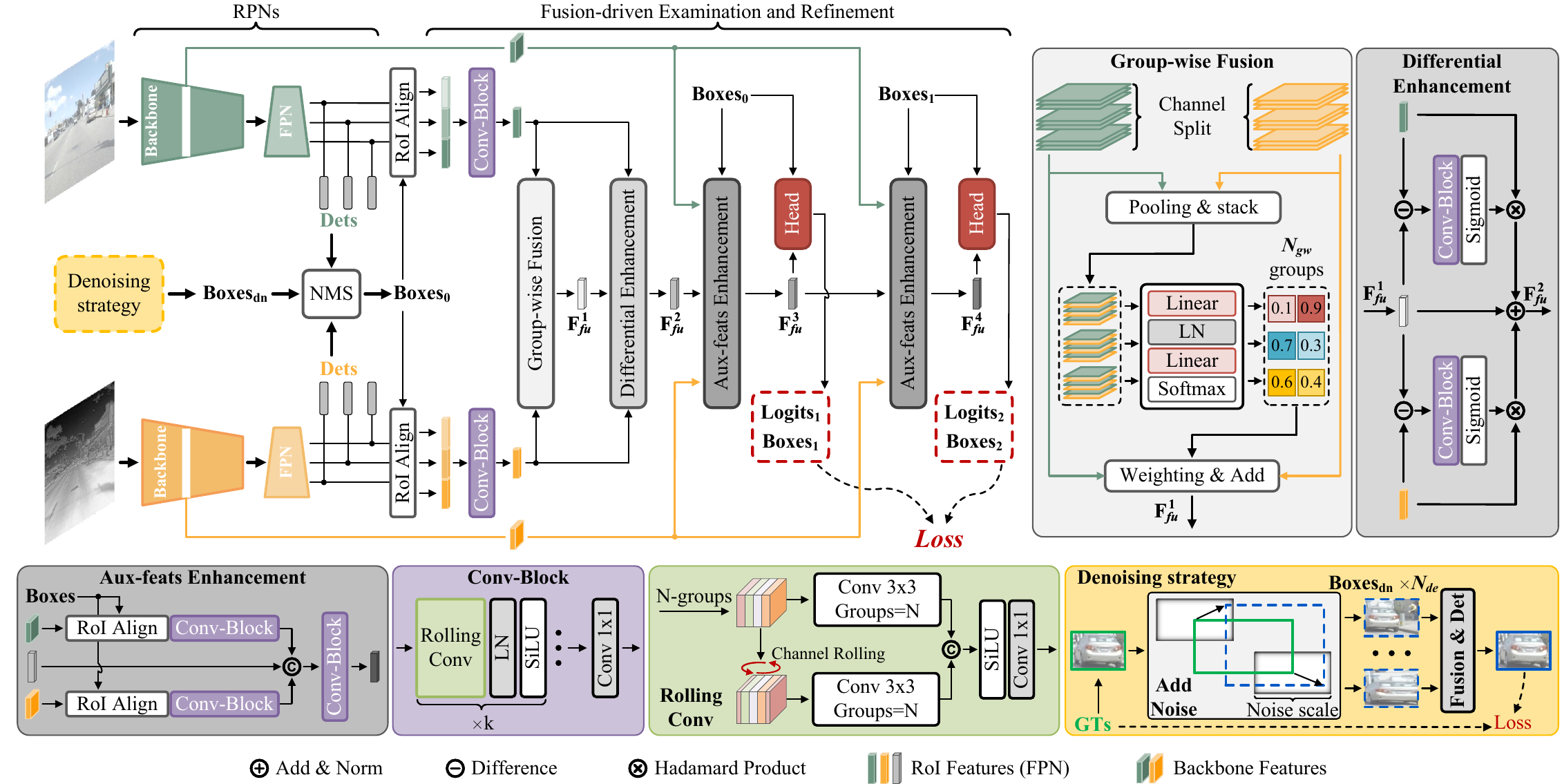}
    \caption{Illustration of the SFEDet framework that adopts the sparse fusion mechanism. It consists of two main components: the dual lightweight RPNs and the elaborate Fusion-driven Examination and Refinement (FER) module. The RPNs filter the background and identify potential foregrounds,~\ie~the RoIs, across the entire image. 
    The FER module performs step-wise cross-modality fusion, classification, and box refinement only within the remaining sparse RoIs.}
    \label{fig:method}
\end{figure}

In this section, we present the instantiation of the sparse fusion mechanism, \ie~the SFEDet framework, which performs efficient and accurate RGB-T detection through two stages: 1) the first stage takes a quick look at the entire image with lightweight model to filter out the redundant backgrounds while retaining potential foreground regions; 2) the second stage performs detailed examination and refinement within the sparse foreground regions, \ie~the RoIs, to produce the final detection results. Next, we describe the overall framework of SFEDet and detail its key components.

\subsection{Overall framework of SFEDet}

Figure~\ref{fig:method} illustrates the overall framework of SFEDet performing detection in two stages. In the first stage, we employ two lightweight detectors serving as independent RPNs to filter out the apparent background regions through preliminary detection. We then construct the proposal set by taking the union set of proposals from individual modalities. 
To further reduce redundancy in the preliminary detections, we subsequently apply NMS in the union set to get sparse, non-overlapping proposals. 
With these proposals, we extract the corresponding RGB-T RoI feature pairs from the multi-level FPN features via RoI Align, where each proposal box is used to get spatially aligned features from two modalities. The proposals generated in the first stage typically contain numerous false positives and poorly localized boxes. To address these issues, we further introduce a Fusion-driven Examination and Refinement (FER) module in the second stage. It fuses the RGB and TIR features through multi-step feature enhancement, producing a more robust representation for each RoI and facilitating the suppression of false positives. For proposals with inaccurate localization, a step-wise refinement strategy is adopted to progressively refine the bounding boxes.

The most computationally intensive component of our SFEDet, \ie~the FER module, is activated only within potential foreground regions. This means that the computational cost depends primarily on the number of RoIs. Consequently, SFEDet exhibits strong scalability when processing high-resolution images. We provide detailed designs of the proposed FER module, the box refinement procedure, and the overall training schedule of SFEDet in the following sections.

\subsection{Fusion-driven Examination and Refinement}
\label{sec:fer}

Due to numerous false positives or poorly localized proposals in the first-stage results, we thus design an elaborate Examination and Refinement module based on the cross-modality feature fusion to address this, as illustrated in Figure~\ref{fig:method}. 
We deploy an efficient differential-enhanced group-wise fusion (DEGF) to balance the fusion cost and quality.
The following multi-step enhancements with auxiliary features ensure a robust representation for each RoI.

\vspace{1mm} \noindent
\textbf{Differential-enhanced group-wise fusion.}
This module contains a group-wise cross-modality fusion and a subsequent differential enhancement module. 
Before fusion, the RoI features from both RGB and TIR modalities are simultaneously extracted from three stages of the feature pyramid network (FPN) via RoI Align, denoted as $\mathbf{F}_{fpn-i}^{M}$, where $M \in \{rgb, tir\}$ and $i \in \{1,2,3\}$. Multi-stage features from FPN are then reduced into a cross-stage representation $\mathbf{F}_{fpn}^{M} \in \mathbb{R}^{C \times 5 \times 5}$ for each proposal, where $C$ denotes the number of channels and 5 represents the RoI spatial size.

Core of this module is the group-weighting fusion applied to $\mathbf{F}_{fpn}^{rgb}$ and $\mathbf{F}_{fpn}^{tir}$, as illustrated in Figure~\ref{fig:method}. After the $N_{gw}$-group split along the channel dimension, the weights within each group are predicted separately and shared across the entire group in fusion, which reduces the computational cost. The softmax normalization ensures that the sum of weights equals one for each channel pair. The fused features are denoted as $\mathbf{F}_{fu}^{1}$.

The differential enhancement module aims to adaptively learn a supplementary representation for each modality by comparing the fused features $\mathbf{F}_{fu}^{1}$ with the original modality-specific features. The adaptive weights are learned from the joint representation of $\mathbf{F}_{fu}^{1}$ and $\mathbf{F}_{fpn}^{M}$, where the feature difference is adopted as a computationally efficient operation. 
Compared with the \textit{Add} and \textit{Concatenation} operation, the adopted difference operation is lighter in computation than the \textit{Concatenation} while achieving superior performance to \textit{Add}, as demonstrated in the ablation study. As illustrated in Figure~\ref{fig:method}, the enhancement procedure is formulated as:
\begin{equation}
\label{eq:diff}
\begin{alignedat}{1}
    \mathbf{\hat{F}}_{rgb} &= \mathbf{F}_{fpn}^{rgb} \cdot \Psi_{rgb}(\mathbf{F}_{fu}^{1} - \mathbf{F}_{fpn}^{rgb}), \\
    \mathbf{\hat{F}}_{tir} &= \mathbf{F}_{fpn}^{tir} \cdot \Psi_{tir}(\mathbf{F}_{fu}^{1} - \mathbf{F}_{fpn}^{tir}), \\
    \mathbf{F}_{fu}^{2} &= \phi_{LN}(\mathbf{F}_{fu}^{1} + \mathbf{\hat{F}}_{rgb} + \mathbf{\hat{F}}_{tir}),
\end{alignedat}
\end{equation}
where $\Psi(\cdot)$ is the convolution block, and $\phi_{LN}$ means the Layer Norm.

\vspace{1mm} \noindent 
\textbf{Auxiliary Feature Enhancement.} 
Since the FPN features are high-level representations extracted from later stages of the network, we introduce low-level features from the backbone (BB) as auxiliary (Aux.) inputs to enhance the fusion representation. The fusion procedure is illustrated in Figure~\ref{fig:method}, where auxiliary features are re-captured by RoI Align from the backbone feature maps.

\vspace{1mm} \noindent
\textbf{Rolling Convolution.}
To further reduce computational costs, we propose a variant of \textit{Group Convolution} (GC), termed Rolling Convolution, which replaces all conventional convolutional operators in the FER module. The performance of cascaded GCs tends to be suboptimal due to the lack of inter-group interactions. To address this issue, we deploy an additional counterpart of GC, where the features are rolled along the channel dimension before convolution. The rolling shift is set to half of the channels within each group, as illustrated in Figure~\ref{fig:method}. A subsequent point-wise convolution merges the outputs from the two GCs. Compared with conventional convolution, the proposed Rolling Convolution employs fewer parameters while providing greater depth and nonlinearity for the detector.

\subsection{Step-wise Box Refinement}
\label{sec:refine}

\textbf{Step-wise Refinement.} 
Owing to the poorly localized bounding boxes in the outputs of lightweight RPNs, we adopt a step-wise refinement strategy to progressively refine the bounding boxes. In the FER module, we employ a two-step refinement scheme, where the bounding boxes are predicted and decoded conditioned on the predictions from the previous step. The residuals between successive bounding boxes are predicted based on the updated fusion features, meaning that the auxiliary features are re-captured via RoI Align at the end of the first step, once the bounding boxes are updated.

\vspace{1mm} \noindent 
\textbf{Box Encoder and Decoder.}
The RPNs in our framework adopt the distance-based decoder to predict bounding boxes for better performance. However, in the FER module, the RoI features lose the stride-scale information, which is inherently preserved in conventional multi-scale feature maps and indicates the actual scale of normalized boxes in the original image space. To address this issue and unify the box constraint format across all heads, we leverage the scale of the proposals to normalize and decode the bounding boxes in the FER module. Let $d_i$ denote the predicted distance of a side, and $L$ denote the theoretical maximum distance that the prediction head can output; the bounding boxes are then decoded as:
\begin{equation}
\label{eq:decode}
D_i= \left\{
\begin{array}{ll}
\frac{x_{2} - x_{1}}{L} \cdot d_i, & i \in \{left, right\}, \\
\frac{y_{2} - y_{1}}{L} \cdot d_i, & i \in \{upper, bottom\},
\end{array}\right.
\end{equation}
where $D_i$ is the updated distance of each side to the center of the proposal box after decoding, and $[x_1, y_1, x_2, y_2]$ indicates the position of the proposal box. The corresponding encoding procedure is then formulated as follows:
\begin{equation}
\label{eq:encode}
\hat{d}_i= \left\{
\begin{array}{ll}
\frac{L}{x_{2} - x_{1}} \cdot D_{i}^{gt}, & i \in \{left, right\}, \\
\frac{L}{y_{2} - y_{1}} \cdot D_{i}^{gt}, & i \in \{upper, bottom\},
\end{array}\right.
\end{equation}
where $D_{i}^{gt}$ is the distance of each side of the ground truth bounding box to the proposal center, $\hat{d}_i$ is the target of the prediction set in the loss function.

\subsection{Denoising Training strategy}
The proposed FER module is trained using the proposals generated by the RPNs. However, during the early stages of training, the RPNs may not be sufficiently robust to cover all regions containing objects. As a result, some regions that potentially contain objects are dropped, which often correspond to hard samples that are crucial for learning. To mitigate this issue, we adopt a denoising training strategy, where $N_{de}$ groups of noisy variants of the ground-truth bounding boxes are added to the proposal list. This strategy enhances the diversity of training samples and encourages the FER module to learn from challenging cases. The procedure of noise injection is illustrated in Figure~\ref{fig:method} and formulated as:
\begin{equation}
\label{eq:denoise}
\begin{alignedat}{1}
    (x_i, y_i) &\rightarrow (x_i +\delta_x, \ y_i + \delta_y), i \in \{1, 2\},  \\
    \delta_x &\sim U(-0.4*W, +0.4*W),   \\
    \delta_y &\sim U(-0.4*H, +0.4*H),
\end{alignedat}
\end{equation}
where \textit{W} and \textit{H} denote the width and height of the GT boxes, respectively. The assigner automatically categorizes these noisy proposals into positive and negative samples and matches them with the corresponding GT boxes.

\subsection{Training Procedure}
\textbf{Two-stage training.}
Our method employs a two-stage training strategy to ensure stable and robust convergence. In the first stage, two independent RPNs are trained separately for the RGB and TIR modalities, and their weights are then initialized in the detector as a warm-up. In the second stage, the entire detector is trained in an end-to-end manner, where the parameter updates of the two RPNs are regulated by a learning rate decay factor~$\lambda$.

\vspace{1mm} \noindent 
\textbf{Total Loss.}
The total loss of our architecture is composed of losses from two independent RPNs and two-step refinements in the FER module. The loss configuration for each head of the FER module can be unified with that of the RPNs based on our encoding scheme proposed in Sec~\ref{sec:refine}. Consequently, all four heads follow the standard loss settings of YOLOv8, where binary cross-entropy (BCE) loss is used for classification, and CIoU and DFL~\cite{li2020generalized} losses are adopted for bounding box prediction. The overall loss function is defined as follows:
\begin{equation} \label{eq:total_loss}
    \mathcal{L}_{total} = \mathcal{L}_{rpn}^{rgb} + \mathcal{L}_{rpn}^{tir} + \mathcal{L}_{fer}^{(1)} + \mathcal{L}_{fer}^{(2)},
\end{equation}
where each loss item is composed of 
\begin{equation}
    \mathcal{L} = \alpha \mathcal{L}_{BCE} + \beta \mathcal{L}_{CIoU} + \gamma L_{DFL}.
\end{equation}
The $\alpha$, $\beta$, and $\gamma$ in Eq~\eqref{eq:total_loss} are the weights of different loss terms.

{
\setlength{\tabcolsep}{2pt}
\begin{table}[!t]
\caption{Comparison with other methods in M3FD benchmark (1024$\times$768). Three-level accent colors (\textcolor{emp_1}{\rule{1ex}{1ex}} \textcolor{emp_2}{\rule{1ex}{1ex}} \textcolor{emp_3}{\rule{1ex}{1ex}}) are respectively used to highlight the top three methods across different metrics, where the \textcolor{emp_1}{red} means the best. CSP means the CSP-Darknet architecture used in YOLOv5/v8. $\dagger$ means our re-evaluated results using the \textit{fvcore} toolkit based on the officially released codes for a fair comparison. 
}
\label{tab:cmp_m3fd}
\centering
{\fontsize{5pt}{6.5pt}\selectfont
\begin{tabularx}{\linewidth}{
lccc
>{\centering\arraybackslash}X
>{\centering\arraybackslash}X
>{\centering\arraybackslash}X
>{\centering\arraybackslash}X
>{\centering\arraybackslash}X
>{\centering\arraybackslash}X
>{\centering\arraybackslash}X
>{\centering\arraybackslash}X
}
\toprule
\multirow{2}[2]{*}{Methods} & \multirow{2}[2]{*}{Backbone}  & \multirow{2}[2]{*}{Params.}    & \multirow{2}[2]{*}{FLOPs} & \multicolumn{6}{c}{Precision (\%)}    & \multirow{2}[2]{*}{mAP}   & \multirow{2}[2]{*}{AP50}  \\ \cmidrule(lr){5-10}
 & & & & People & Bus   & Car   & Motor & Lamp  & Truck & & \\ \midrule 

(2024') ICAFusion~\cite{shen2024icafusion}  & CSP-L & 120M  & 370G$^\dagger$ & 82.3  & 93.9  & 93.4  & 89.6  & 88.2  & 87.8  & --    & \textcolor{emp_2}{88.2}  \\
(2024') EME~\cite{Rethink_TIV}  & Res-50 & \textcolor{emp_2}{36.4M} & \textcolor{emp_2}{316G$^\dagger$}    & 79.5  & 89.8  & 91.9  & 74.87 & 77.4  & 84.0  & 54.0  & 82.9  \\
(2025') RDMI~\cite{Tian_Yang_Zhu_Wang_He_2025}  & CSP-L  & \textcolor{emp_3}{72.0M}   & \textcolor{emp_3}{320G}  & 84.3  & 85.4  & 87.7  & 68  & 79.6  & 70.4  & 49.7  & 79.2 \\
(2025') MMFN~\cite{Yang_Liang_Li_Zhang_2025}    & Res-50 & 176M  & --    & 83.0  & 92.1  & 93.2  & 73.7  & 87.6  & 87.4  & --    & 86.2  \\
(2025') EI2Det~\cite{Hu_Edge_2025}   & CSP-L & 116M   & 391G$^\dagger$  & 82.8  & 91.6  & 91.4  & 77.2  & 84.7  & 89.3  & \textcolor{emp_3}{55.5}  &86.2   \\
(2025') Fu-Mamba~\cite{FusionMamba}  & CSP-L & 288M    & 1133G$^\dagger$ & 84.3 & 94.2 & 92.9  & 80.5  & 87.5  & 88.8  & \textcolor{emp_1}{61.9}  & \textcolor{emp_3}{88.0}  \\
\midrule
SFEDet (Ours) & CSP-S & \textcolor{emp_1}{24.5M} & \textcolor{emp_1}{112.6G} & 88.7  & 96.7  & 91.2  & 88.2  & 91    & 83.2  & \textcolor{emp_2}{61.0}    & \textcolor{emp_1}{89.8}  \\
\bottomrule
\end{tabularx}
}
\end{table}
}

\section{Experiments}

\subsection{Experimental Setup}
\textbf{Datasets.} M3FD~\cite{m3fd} is a well-aligned RGB-T dataset widely used in RGB-T object detection and image fusion. It contains 6 categories and is built with 4.2K image pairs at 1024$\times$768 resolution. The dataset contains several challenging scenarios with extreme light and weather conditions, and objects of different scales, making it hard for any single modality. The relabeled FLIR~\cite{zhang2020multispectral_flir} benchmark is a large-scale dataset widely adopted in RGB-T object detection, including 3 categories at 640$\times$512 resolution. The LLVIP~\cite{llvip} is a high-quality RGB-T pedestrian detection dataset with more than 15K image pairs at 1280$\times$1024 resolution. 

\vspace{1mm} \noindent 
\textbf{Evaluation metrics.} We use the standard COCO-style
\textit{mAP 0.5:0.95} (denoted as \textit{mAP}) and \textit{AP50} as the detection metric throughout all experiments. 

\vspace{1mm} \noindent 
\textbf{Implementation details.} We deploy the YOLOv8-Small as the RPN in all experiments. The $N_{gw}$ in group-wise fusion is 4, and the $N_{de}$ in denoising training is 10 to balance the training cost. The $L$ in the box-decoding is 16, as that in standard YOLOv8. The groups in the rolling convolution and the RoI size are detailed in the ablation study. The weights of loss items are set to $\alpha=0.5$, $\beta=7.5$, and $\gamma=0.375$.
We adopt the AdamW optimizer for training, with a learning rate of 0.0002 and a batch size of 6. The learning rate decay for the RPNs, \ie~the $\lambda$, is set to 0.005.
Please refer to \textbf{Appendix B} for more details.

{
\setlength{\tabcolsep}{2.1pt}
\begin{table}[!t]
\caption{Comparison with other methods in FLIR and LLVIP benchmarks. SFEDet-Lite represents a sparser configuration of our method, with the decrease in AP50 kept within 0.1. Three-level accent colors (\textcolor{emp_1}{\rule{1ex}{1ex}} \textcolor{emp_2}{\rule{1ex}{1ex}} \textcolor{emp_3}{\rule{1ex}{1ex}}) are respectively used to highlight the top three methods across different metrics, where the \textcolor{emp_1}{red} means the best. $\dagger$ means our re-evaluated results based on the officially released codes. 
}
\label{tab:cmp_flir}
\centering
{\fontsize{5pt}{6.5pt}\selectfont
\begin{tabularx}{\linewidth}{
lccc
>{\centering\arraybackslash}X
>{\centering\arraybackslash}X
>{\centering\arraybackslash}X
c
c
c
c
c
c
}
\toprule
\multirow{2}[2]{*}{Methods} & \multirow{2}[2]{*}{Backbone}  & \multirow{2}[2]{*}{Params.}    & \multirow{2}[2]{*}{FLOPs} & \multicolumn{3}{c}{Precision (\%)}    & \multirow{2}[2]{*}{mAP}   & \multirow{2}[2]{*}{AP50} & \multirow{2}{*}{} & \multirow{2}[2]{*}{FLOPS} & \multirow{2}[2]{*}{mAP}   & \multirow{2}[2]{*}{AP50} \\ \cmidrule(lr){5-7}

 & & & & Bicycle    & Car   & Person    & & & & & &    \\ \midrule
 \multicolumn{9}{c}{\cellcolor[HTML]{EFEFEF} FLIR (640$\times$512)} & &\multicolumn{3}{c}{\cellcolor[HTML]{EFEFEF} LLVIP (1280$\times$1024)} \\
(2021') CFT~\cite{cft}  & CSP-L  & 206M  & 224G   & --    & --    & --    & 40.2  & 78.7  & & 640G$^\dagger$     & 63.6    & \textcolor{emp_3}{97.5} \\
(2024') ICAFusion~\cite{shen2024icafusion}  & CSP-L  & 120M  & 166G$^\dagger$ & 66.9  & 89.0 & 81.6  & 41.1  & 79.2 & & -- & -- & -- \\
(2024') Cro-Former~\cite{lee2024crossformer}   & Res-50 & --    & 362G  & --    & --    & --    & 42.1  & 79.3  & & -- & 65.1  & 97.4  \\
(2025') RDMI~\cite{Tian_Yang_Zhu_Wang_He_2025}  & CSP-L  & \textcolor{emp_3}{72M}   & \textcolor{emp_3}{133G} & 62.0    & 89.3  & 85    & 41.2  & 78.8  & & -- & -- & --    \\
(2025') MMFN~\cite{Yang_Liang_Li_Zhang_2025}    & Res-50 & 176M  & --    & 65.5  & 91.2  & 85.7  & 41.7  & 80.8 &  & -- & --  & 97.2  \\
(2025') COFNet~\cite{Zhou_Li_Yang_Wei_Pu_Luo_Jia_2025}  & CSP-L  & 90.2M & 197G  & --    & --    & --    & \textcolor{emp_2}{44.6}  & \textcolor{emp_2}{83.6} & & \textcolor{emp_3}{628G}$^\dagger$ & \textcolor{emp_1}{65.9}    & \textcolor{emp_2}{97.7}  \\
(2025') EI2Det~\cite{Hu_Edge_2025}   & CSP-L  & 116M   & 168G$^\dagger$  & 66.3  & 89.4  & 84.9  & --    & 80.2 & & 644G$^\dagger$ & 63.9    & \textcolor{emp_1}{98.0}  \\
(2025') Fu-Mamba~\cite{FusionMamba}  & CSP-L  & 288M    & 472G$^\dagger$     & --    & --    & --    & \textcolor{emp_1}{45.9}  & \textcolor{emp_1}{84.9} & & 1888G$^\dagger$ & 64.3  & 97.0 \\    
\midrule
SFEDet (Ours) & CSP-S & \textcolor{emp_2}{24.5M} & \textcolor{emp_2}{69.0G} & 70.6  & 89.3  & 85.2    & \textcolor{emp_3}{43.0}  & \textcolor{emp_3}{81.7} & & \textcolor{emp_2}{163G} & \textcolor{emp_2}{65.9}    & 96.8 \\
SFEDet-Lite (Ours) & CSP-S & \textcolor{emp_1}{24.5M} & \textcolor{emp_1}{53.6G}    & 70.5  & 89.1  & 85.2  & 42.8  & 81.6  & & \textcolor{emp_1}{153G}   & \textcolor{emp_3}{65.6}   & 96.8     \\  
\bottomrule
\end{tabularx}
}
\end{table}
}

\subsection{Comparisons with State-of-the-art Methods}

\textbf{Results on M3FD dataset.} 
The quantitative comparison on the M3FD dataset is presented in Table~\ref{tab:cmp_m3fd}. The term \textit{CSP} in the table refers to the CSP-Darknet used in the YOLOv5/v8 series. We re-evaluated several methods using their official code to measure their computational cost at a resolution of 1024×768, which are marked with $^\dagger$. Our method achieves competitive detection accuracy while requiring only about 30\% of the computational cost compared to other approaches on the high-resolution images.


\begin{figure}[!t]
\centering
\includegraphics[width=0.55\linewidth]{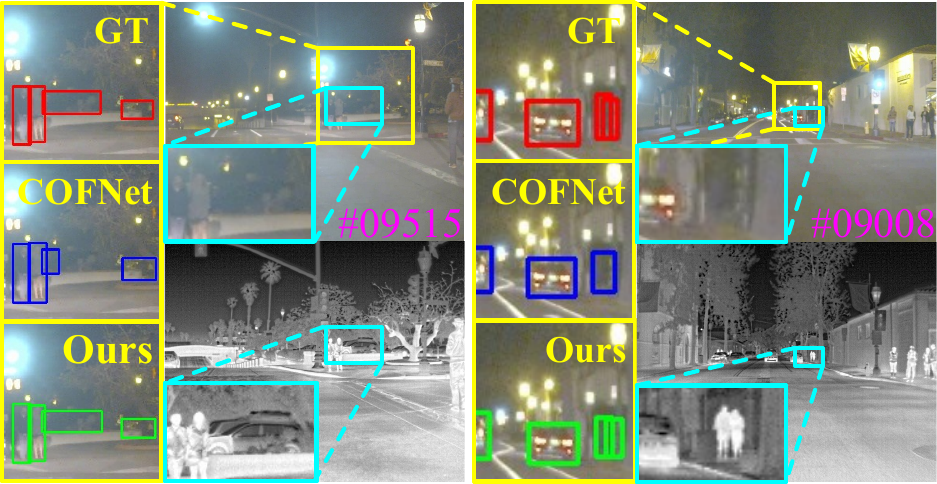} 
\caption{Qualitative comparison with COFNet in challenging illumination conditions, demonstrating the effectiveness of our SFEDet. }
\label{fig:challeng}
\end{figure}

\vspace{1mm} \noindent  
\textbf{Results on FLIR and LLVIP datasets.} 
Table~\ref{tab:cmp_flir} presents comparison results on them. The proposed SFEdet achieves precision comparable to significantly heavier detectors. On the LLVIP dataset, the image area increases by 400\%, leading to a substantial rise in computational cost. SFEDet experiences a 2.4$\times$ increase in FLOPs, which is smaller than that of other methods, for example, 3.8$\times$ in EI2Det, demonstrating the advantage of our method in maintaining sub-linear growth in computational complexity. Figure~\ref{fig:challeng} presents a qualitative comparison between COFNet and our SFEDet in challenging illumination conditions, demonstrating the effectiveness of our sparse fusion architecture. 
More qualitative detection results on different datasets are in \textbf{Appendix B}.

Furthermore, we construct a lighter SFEDet version for the FLIR and LLVIP datasets by filtering proposals and increasing sparsity, denoted as SFEDet-Lite. We set the configuration to limit the AP50 decrease within 0.1, presented in Table~\ref{tab:cmp_flir}, which achieves an additional reduction of approximately 10G FLOPs.

{
\setlength{\tabcolsep}{4pt}
\begin{table}[!t]
\centering
\caption{Comparison with detectors via dense fusion in the same basic setting. The best and second-best methods are indicated in bold and underlined, respectively. 
}
\label{tab:other_fusion}
{\fontsize{5pt}{6.5pt}\selectfont
\begin{tabularx}{0.65\linewidth}{
llc
>{\centering\arraybackslash}X
>{\centering\arraybackslash}X
c}
\toprule
Fusion after & Backbone  & Params.   & mAP   & AP50  & FLOPs \\ \midrule
\multicolumn{6}{c}{\cellcolor[HTML]{EFEFEF} M3FD (1024$\times$768)} \\
Image       & 1$\times$CSP-L    & 51.2M & \textbf{62.5}  & \textbf{89.8}  & 268G   \\
Backbone    & 2$\times$CSP-S    & 18.8M & 55.6  & 85.5  & 88.0G    \\    
FPN (+Our DEGF)    & 2$\times$CSP-S & 18.8M & 59.6 & 88.4  & 180.4G \\
\rowcolor[HTML]{DBEEF9} 
SFEDet (Ours) & 2$\times$CSP-S  & 24.5M & \underline{61.0}    & \underline{89.8}  & 112.6G     \\  \midrule
\multicolumn{6}{c}{\cellcolor[HTML]{EFEFEF} FLIR (640$\times$512)} \\
Image       & 1$\times$CSP-L    & 51.2M & 39.5  & 77.6  & 130.0G       \\
Backbone    & 2$\times$CSP-S    & 18.8M & 38.2  & 76.5  & 36.8G     \\
FPN (+Our DEGF)    & 2$\times$CSP-S & 18.8M & \underline{41.4} & \underline{79.2}  & 75.2G \\
\rowcolor[HTML]{DBEEF9} 
SFEDet (Ours) & 2$\times$CSP-S  & 24.5M & \textbf{43.0}  & \textbf{81.7}  & 69.0G     \\
\bottomrule
\end{tabularx}
}
\end{table}
}

{
\setlength{\tabcolsep}{2pt}
\begin{table*}[!t]
\caption{Ablation study on the design of each component. BB denotes the backbone; the first line is the vanilla setting, and blue highlights the setting we finally adopt. 
}
\label{tab:ablation}
\centering

{\fontsize{5pt}{6.5pt}\selectfont
\begin{tabularx}{\linewidth}{
c
>{\centering\arraybackslash}X
>{\centering\arraybackslash}X
>{\centering\arraybackslash}X
>{\centering\arraybackslash}X
>{\centering\arraybackslash}X
>{\centering\arraybackslash}X
ccccccc
}
\toprule
& & & & & & & & \multicolumn{3}{c}{M3FD (1024$\times$768)}  & \multicolumn{3}{c}{FLIR (640$\times$512)}   \\ \cmidrule(lr){9-11} \cmidrule(lr){12-14}
& \multirow{-2}{*}[2pt]{\makecell{GW \\ Fu.}} & \multirow{-2}{*}[2pt]{Diff.} & \multirow{-2}{*}[2pt]{\makecell{Aux-\\ feats}} & \multirow{-2}{*}[2pt]{\makecell{2nd- \\ Ref.}} & \multirow{-2}{*}[2pt]{\makecell{Roll. \\ Conv}} & \multirow{-2}{*}[2pt]{DN} & \multirow{-2}{*}[2pt]{\makecell{Params. \\ (M)}} & mAP  & AP50 & FLOPs (G)                & mAP  & AP50 & FLOPs (G)                \\ \midrule

Base & \textcolor{black!30}{\ding{56}} & Add   &  \textcolor{black!30}{\ding{56}}   & \textcolor{black!30}{\ding{56}}     & \textcolor{black!30}{\ding{56}}  & \textcolor{black!30}{\ding{56}}  & 21.1                       & 57.2 & 85.4 & 104.0       & 41.9 & 79.9 & 58.0 \\ \cmidrule(){1-14}
& \ding{52} & Add   &  \textcolor{black!30}{\ding{56}}   & \textcolor{black!30}{\ding{56}}     & \ding{52}  & \ding{52}  & 20.0                       & 58.4 & 87.1 & 96.4       & 41.6 & 79.9 & 45.8 \\
& \textcolor{black!30}{\ding{56}} & \ding{52} & \textcolor{black!30}{\ding{56}} & \textcolor{black!30}{\ding{56}}  & \ding{52} & \ding{52} & 20.0   & 58.1  & 87.4  & 96.4  & 41.7  & 79.7  & 45.8 \\
& \ding{52} & \ding{52}  & \textcolor{black!30}{\ding{56}} & \textcolor{black!30}{\ding{56}} & \ding{52}  & \ding{52}     & 20.0                      & 58.6 & 87.2 & 96.4       & 41.9 & 80.3 & 45.8       \\
& \ding{52} & \ding{52}    & RPN  & \textcolor{black!30}{\ding{56}} & \ding{52}  & \ding{52}   & 21.1      & 59.1 & 88.1 & 104.2    & 41.9 & 79.9 & 57.2       \\
& \ding{52} & \ding{52}  & BB   &  \textcolor{black!30}{\ding{56}}   & \ding{52}    & \ding{52}   & 21.1                   & 59.9 & 88.5 & 104.6    & 42.1 & 80.8 & 57.0       \\
\rowcolor[HTML]{DBEEF9} FER & \ding{52} & \ding{52} & BB     & \ding{52}   & \ding{52}   & \ding{52}   & 24.5   & 61.0   & 89.8 & 112.6    & 43.0   & 81.7 & 69.0       \\
& \ding{52} & \ding{52}   & BB   & \ding{52}   & \textcolor{black!30}{\ding{56}}  & \ding{52}   & 29.0   & 61.1 & 88.9 & 144.4    & 42.8 & 81.2 & 116.0   \\
& \ding{52} & \ding{52}   & BB    & \ding{52}  & \ding{52}    & \textcolor{black!30}{\ding{56}} & 24.5   & 59.9 & 88.0   & 111.6    & 42.7 & 81.0  & 68.4        \\
\bottomrule
\end{tabularx} 
}
\end{table*}
}

\subsection{Ablation Study} 

\textbf{Comparison with other fusion manners.}
To quantitatively analyze the performance and complexity of our sparse fusion detection framework, we construct several RGB-T detector prototypes using different fusion strategies, as summarized in Table~\ref{tab:other_fusion}. We perform dense fusion on image pairs, backbone features, and FPN features via channel-wise concatenation. Given the relatively limited performance of the image-level fusion detector, we deploy the \textit{Large} model for comparison. Furthermore, we apply the proposed DEGF module in a dense fusion manner to further demonstrate the efficiency of the sparse fusion mechanism. Experimental results on the M3FD and FLIR datasets indicate that our method achieves a favorable trade-off between computational cost and performance.

\vspace{1mm} \noindent 
\textbf{Effects of model components.} 
The ablation study of different modules is summarized in Table~\ref{tab:ablation}. The results indicate that the group-wise fusion provides better performance due to fine-grained adaptation in fusion, without any increased cost. Compared with using element-wise addition, the differential enhancement module improves precision without increasing computational cost. The auxiliary feature column demonstrates the necessity of the third-stage feature enhancement, where low-level features extracted from the backbone are more effective than high-level features from the FPN. The second-refinement column illustrates the improvement introduced by the step-wise refinement mechanism. All convolutional layers in the FER module are replaced by rolling convolution layers, which reduce computational complexity while enhancing performance. This may be attributed to the increased depth of the network. Without the denoising training strategy, the detector performance declines, possibly because the model lacks sufficient hard training samples.

{
\setlength{\tabcolsep}{4pt}
\begin{table}[!t]
\centering
\caption{Comparison of different convolution operations. The FLOPs column reports the statistics of computational costs at the 25\%, 50\%, and 75\% percentiles, as well as their mean values.}
\label{tab:conv}
{\fontsize{5pt}{6.5pt}\selectfont
\begin{tabularx}{0.65\linewidth}{
l
c
>{\centering\arraybackslash}X
>{\centering\arraybackslash}X
c
}
\toprule
Conv OP (dim=128)       & Params. & mAP  & AP50 & FLOPs (G)                \\ \midrule
\multicolumn{5}{c}{\cellcolor[HTML]{EFEFEF} M3FD (1024$\times$768)}       \\ \rowcolor[HTML]{DBEEF9} 
Rolling (G=8)  & 24.5M   & \underline{61.0}   & \textbf{89.8} & {[}92, 110, 122{]} -- 112.6  \\
Shuffle~\cite{shufflenet} (G=8)  & 22.6M   & 59.8 & 88.9 & {[}90, 98, 102{]} -- 98.4  \\
DW + Conv 1$\times$1 & 22.6M   & 59.6 & 88.5 & {[}90, 98, 102{]} -- 98.4  \\
Conv2d        & 29.0M     & \textbf{61.1} & \underline{88.9} & {[}96, 144, 168{]} -- 144.4  \\ \midrule
\multicolumn{5}{c}{\cellcolor[HTML]{EFEFEF} FLIR (640$\times$512)}        \\ \rowcolor[HTML]{DBEEF9} 
Rolling (G=8)  & 24.5M   & \textbf{43.0} & \textbf{81.7} & {[}64, 68, 72{]} -- 69.0  \\
Shuffle~\cite{shufflenet} (G=8)  & 22.6M   & 42.6 & 80.9 & {[}46, 48, 50{]} -- 49.0  \\
DW + Conv 1$\times$1 & 22.6M   & 42.4 & 80.6 & {[}46, 48, 50{]} -- 49.0 \\
Conv2d        & 29.0M     & \underline{42.8} & \underline{81.2} & {[}102, 114, 124{]} -- 116.0  \\ \bottomrule 
\end{tabularx}
}
\end{table}
}

\vspace{1mm} \noindent 
\textbf{Effects of different convolution operations.}
Lightweight variants of the conventional convolutional layer have been widely studied, including group convolution, depthwise (DW) convolution, and the ShuffleNet~\cite{shufflenet} block. We perform a comparative study among several existing lightweight convolution designs, as summarized in Table~\ref{tab:conv}. The results indicate that all lightweight convolutions achieve performance comparable to other detectors, while the proposed rolling convolution further improves performance at slightly higher computational cost.

\begin{figure}[!t]
\centering
\includegraphics[width=0.55\linewidth]{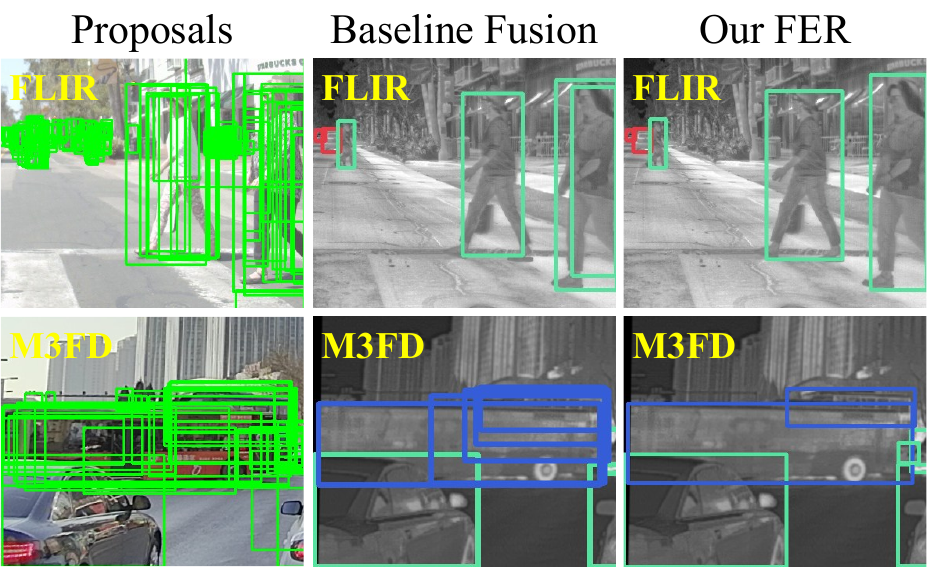} 
\caption{Qualitative comparison between FER and vanilla fusion.}
\label{fig:qualitative_fer}
\end{figure}

{
\setlength{\tabcolsep}{3.2pt}
\begin{table}[!t]
\centering
\caption{The AP50 at different proposal localization accuracy.}
\label{tab:ppl_iou}
{\fontsize{5pt}{6.5pt}\selectfont
\begin{tabularx}{0.65\linewidth}{
l
>{\centering\arraybackslash}X
>{\centering\arraybackslash}X
>{\centering\arraybackslash}X
>{\centering\arraybackslash}X
>{\centering\arraybackslash}X
>{\centering\arraybackslash}X|
c
}
\toprule
\multirow{2}[2]{*}{Data} & \multicolumn{6}{c|}{Proposal IoU Upper Bound} & \multirow{2}[2]{*}{\makecell{EI2Det \\ (2025')}} \\ \cmidrule(lr){2-7}
 & 1.0 & 0.9 & 0.8 & 0.7 & 0.6 & 0.5 &  \\ \midrule
FLIR & \cellcolor{green!20}81.7 & \cellcolor{green!20}81.6 & \cellcolor{green!20}81.7 & \cellcolor{green!20} 81.5 & 79.3 & 62.7 & 80.2 \\
M3FD & \cellcolor{green!20}89.8 & \cellcolor{green!20}89.8 & \cellcolor{green!20}89.8 & \cellcolor{green!20}89.5 & \cellcolor{green!20}88.2 & 75.3 & 86.2 \\
\bottomrule
\end{tabularx} 
}
\end{table}
}

{
\setlength{\tabcolsep}{4pt}
\begin{table}[!t]
\centering
\caption{Ablation study on hyper-parameters in our framework. The FLOPs column reports the statistics of computational costs at the 25\%, 50\%, and 75\% percentiles, as well as their mean values.}
\label{tab:hyper}
{\fontsize{5pt}{6.5pt}\selectfont
\begin{tabularx}{0.65\linewidth}{
cc
>{\centering\arraybackslash}X
>{\centering\arraybackslash}X
cc}
\toprule
RoI-size & RC-group & mAP  & AP50 & Params. & FLOPs (G)              \\ \midrule
7        & 8        & 61.0 & 88.9 & 26.8M   & {[}96, 128, 154{]} -- 134.0 \\ \rowcolor[HTML]{DBEEF9} 
5        & 8        & 61.0 & 89.8 & 24.5M   & {[}92, 110, 122{]} -- 112.6 \\
3        & 8        & 60.0 & 88.7 & 22.9M   & {[}90, 98, 102{]} -- 98.8 \\
5        & 4        & 61.1 & 89.7 & 25.7M   & {[}94, 120, 134{]} -- 121.6 \\
5        & 16       & 60.8 & 88.9 & 23.9M   & {[}92, 106, 114{]} -- 107.8 \\ \bottomrule
\end{tabularx} 
}
\end{table}
}

\vspace{1mm} \noindent 
\textbf{Total Effects of FER.}
The experimental results demonstrate the effectiveness of FER in improving localization quality, as shown in Table~\ref{tab:ablation}. The mAP of the baseline increases largely in the FLIR and M3FD. The qualitative experiment on the typical conditions shown in Figure~\ref{fig:qualitative_fer} demonstrates the effectiveness of FER. The quantitative results show that the false-positive count is reduced from 26.5 to 2.1 per image on M3FD. More details are in \textbf{Appendix A}. 

To evaluate the robustness of FER on poor localized proposals, we progressively noise the proposals to ensure that their IoUs are below the predefined upper bound  (from 1.0 to 0.5) to simulate providing poor proposals for FER. The corresponding results are reported in Table~\ref{tab:ppl_iou}, where all results are obtained \textit{without retraining}, and demonstrate the robustness of our SFEDet. This robustness owes to 2 designs of SFEDet: 1) the proposed denoising training strategy; and 2) the scale factor of 1.1 to enlarge the proposals before pooling. 

\vspace{1mm} \noindent 
\textbf{RoI size and Rolling Convolution groups.}
The feature size obtained from RoI Align and the group configuration in the rolling convolution are two key hyperparameters in our method, as they correspond to detection performance and computational complexity. We conduct an additional ablation study to investigate their effects, as summarized in Table~\ref{tab:hyper}. Configurations highlighted in blue represent the balanced setting that achieves a desirable trade-off between cost and performance and are thus adopted in the final model configuration.

\begin{figure}[!t]
    \centering
    \includegraphics[width=0.32\linewidth]{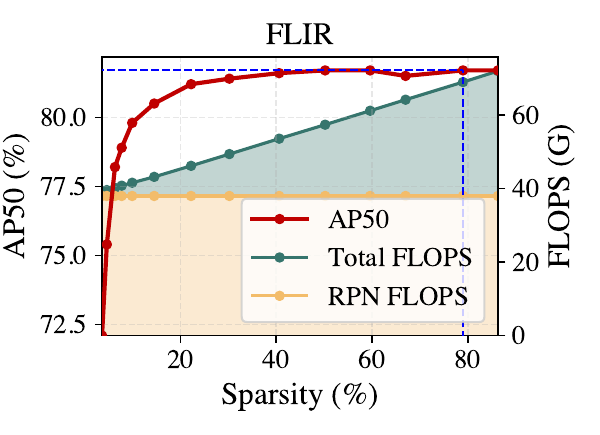} 
    \includegraphics[width=0.32\linewidth]{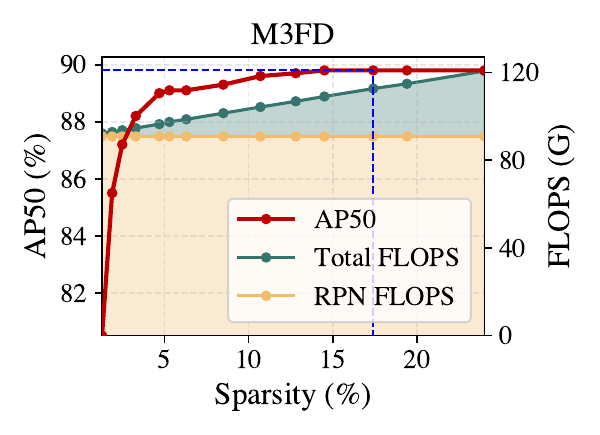} 
    \includegraphics[width=0.32\linewidth]{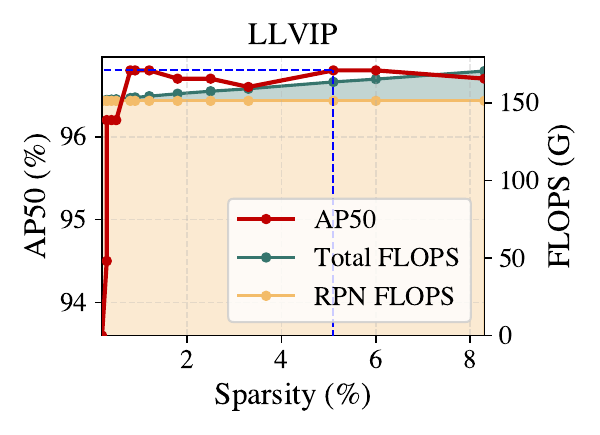} 
    \caption{The relationship between precision/complexity and sparsity. The results on different benchmarks demonstrate that our detector maintains high precision even under large sparsity, while significantly reducing computational cost, especially for high-resolution images. Blue lines indicate the setting of our reported performance.} 
    \label{fig:sparse}
\end{figure}

{
\begin{figure}[!t]
\centering
\begin{minipage}[b]{0.485\linewidth}
\centering
\includegraphics[width=\linewidth]{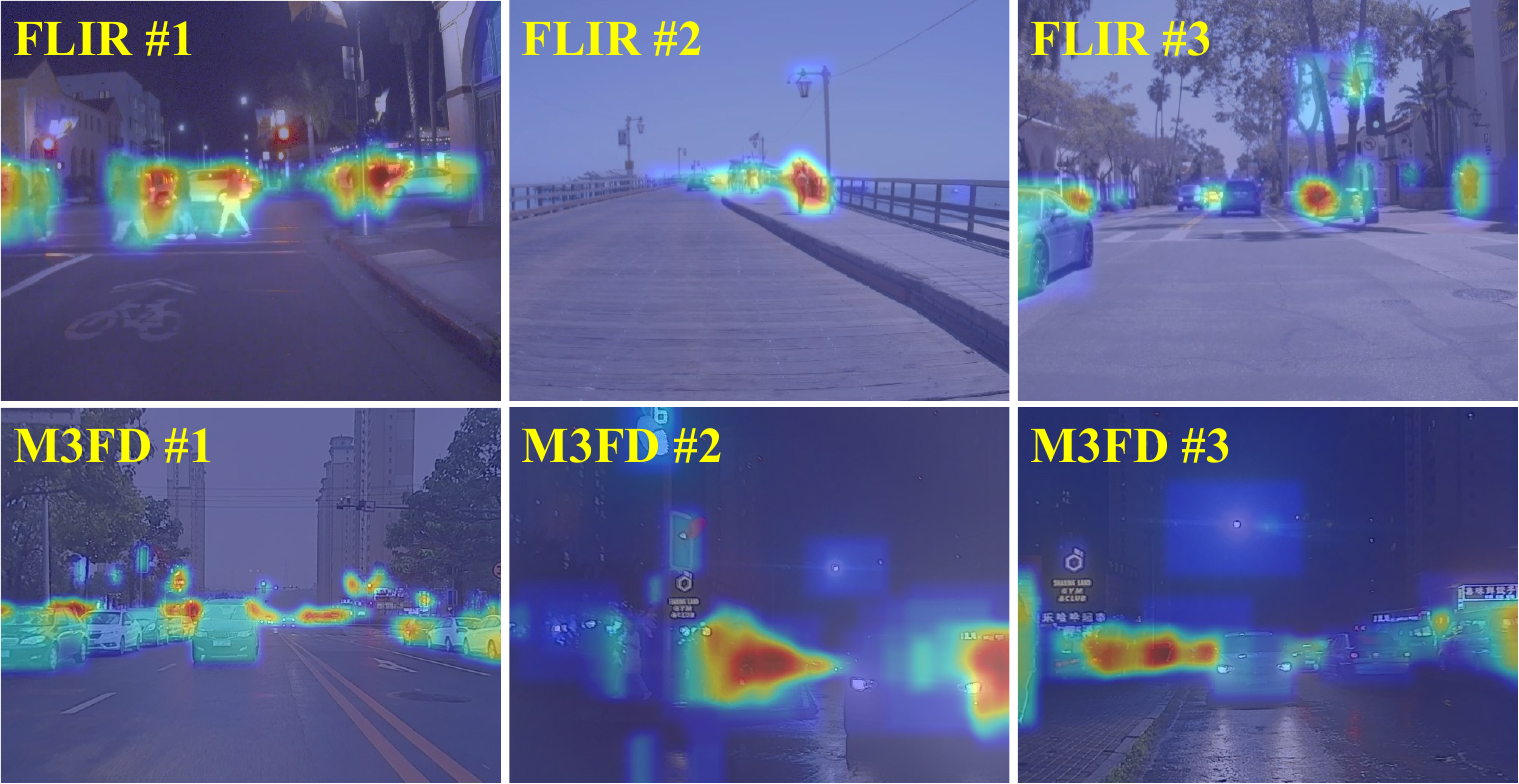} 
\captionof{figure}{Complexity distribution map. The red areas are allocated with more FLOPs, while the blue areas are allocated with the minimum computational cost. The apparent backgrounds are filtered out, and the fusion is performed on the potential areas.} 
\label{fig:sparse_map}
\end{minipage} \hfill \begin{minipage}[b]{0.475\linewidth}
\centering
\includegraphics[width=0.75\linewidth]{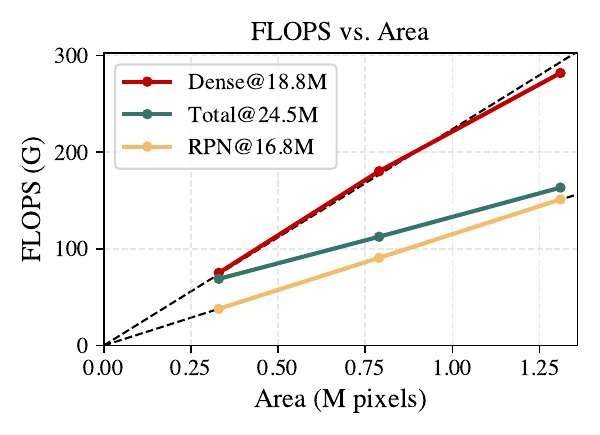} 
\captionof{figure}{Computational cost vs. image area. The \textcolor[HTML]{C00000}{Dense} is the FPN-feature fusion detector we implement with DEGF in Table~\ref{tab:other_fusion}. \textcolor[HTML]{f3bc6a}{RPN} means the total of two RPN modules in our framework. The \textcolor[HTML]{36756d}{Total} line means our SFEDet.} 
\label{fig:floaps_area}
\end{minipage} 
\end{figure}
}

\subsection{Complexity Analysis}

\textbf{FLOPs \textit{vs.} Sparsity.}
The low computational cost of our SFEDet primarily results from its sparse fusion and prediction mechanisms. Consequently, the prediction sparsity has a significant impact on the overall computational cost. We further investigate the relationship between FLOPs and prediction sparsity, which we define as: 
\begin{equation} 
Sparsity = n \cdot (S_{roi})^2 \ / \sum_{level=1}^{3} w_{fpn}^{level} \cdot h_{fpn}^{level}, 
\end{equation}
where $n$ denotes the number of proposals, $S_{roi}$ is the RoI feature size, and $w_{fpn}$ and $h_{fpn}$ represent the feature map widths and heights at different levels, respectively. This metric represents the ratio of the spatial area where convolution-based computations are performed. We adjust the sparsity by varying the threshold of RPN outputs, as illustrated in Figure~\ref{fig:sparse}. The curves indicate that the computational cost introduced by the FER is linearly correlated with the number of proposals. Experimental results on different datasets suggest that our detector maintains high precision even under large sparsity, while substantially reducing the overall computational cost. In high-resolution images, most of the computation is dominated by the RPN, whereas the FER accounts for only a small fraction of the total cost. The qualitative results in Figure~\ref{fig:sparse_map} visualize the spatial distribution of computational complexity, showing that the computation is sparse with most background regions excluded.

\vspace{1mm} \noindent  
\textbf{FLOPs \textit{vs.} Image size.}
We further examine the relationship between computational complexity and image area within our framework. We respectively evaluate the RPN module, SFEDet, and its dense variant with DEGF on datasets of varying resolutions. Quantitative results are presented in Figure~\ref{fig:floaps_area}. The RPN module and SFEDet-Dense are fully convolutional networks, and their computational costs scale linearly with the detection area. Our SFEDet achieves improved detection performance with only a modest increase in FLOPs, and exhibits sub-linear growth in computational cost with respect to detection area.

{
\begin{figure}[!t]
\centering
\begin{minipage}[t]{0.31\linewidth}
\centering
\includegraphics[width=\linewidth]{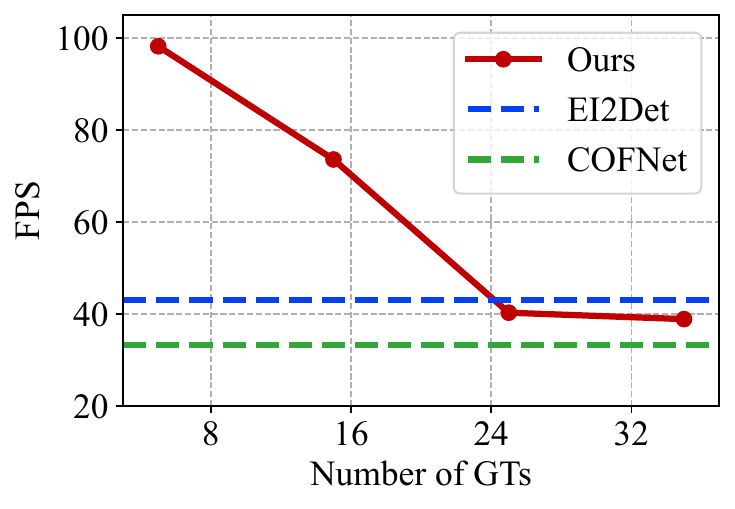} 
\captionof{figure}{FPS vs. \#GTs. SFEDet achieves comparable efficiency on dense samples and much better on sparse samples.} 
\label{fig:gt_fps}
\end{minipage} \hfill \begin{minipage}[t]{0.632\linewidth}
\centering
\includegraphics[width=0.49\linewidth]{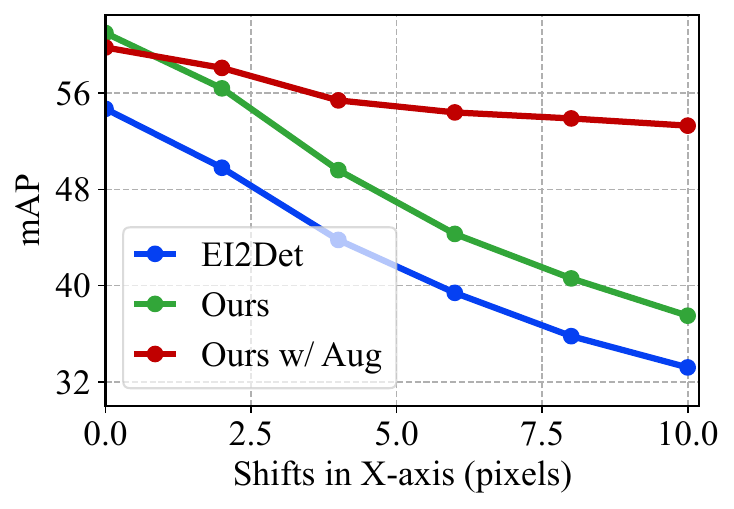} 
\includegraphics[width=0.49\linewidth]{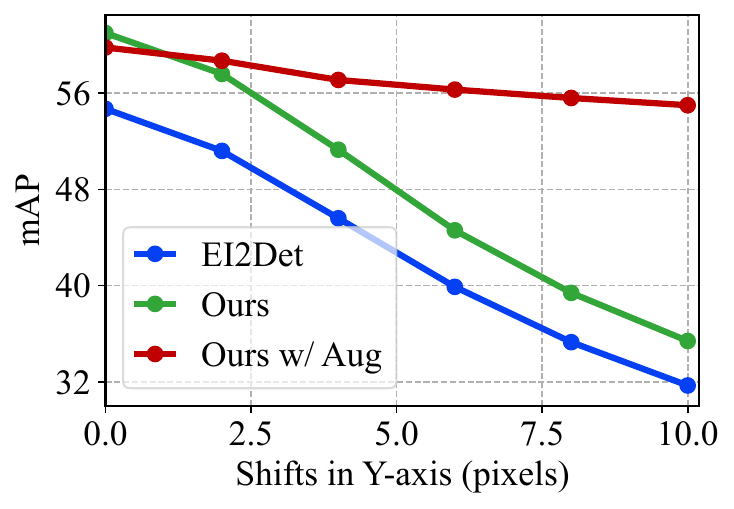}
\captionof{figure}{mAP vs. spatial shifts. Spatial misalignment degrades the performance of both our method and the dense fusion detector EI2Det. The impact of misalignment can be effectively alleviated by "shift" data augmentation in training. }
\label{fig:shifts}
\end{minipage}
\end{figure}
}

\vspace{1mm} \noindent  
\textbf{FPS \textit{vs.} Number of GTs.} The FPS of our SFEDet is related to the number of RoIs, \ie~the objects. The quantitative evaluation results are shown in Figure~\ref{fig:gt_fps}. Our method achieves a comparable FPS to those large dense detectors~\cite{Zhou_Li_Yang_Wei_Pu_Luo_Jia_2025,Hu_Edge_2025}, while largely outperforming them in samples with fewer objects.

\subsection{Discussion on spatial misalignment}

The spatial shifts between the RGB and TIR images lead to a misalignment in feature fusion, which impacts both the dense fusion detector (\eg~EI2Det~\cite{Hu_Edge_2025}) and our SFEDet, as shown in Figure~\ref{fig:shifts}. In recent research~\cite{tian2024cross}, some specialized techniques are developed to handle this issue. We adopt the corresponding data augmentation, \ie~the shift, in the training of our SFEDet. The impact of misalignment can be effectively alleviated, as shown by the red line in Figure~\ref{fig:shifts}, demonstrating the practical effectiveness of our sparse fusion architecture.

\section{Conclusion}

In this work, we propose an efficient RGB-T detector, SFEDet. By filtering most background at the very beginning, the SFEDet performs feature fusion-driven examination and step-wise refinements via the proposed FER module only within potential RoIs, thereby improving the computational efficiency. 
We introduce a Rolling Convolution operator and a denoising training strategy to reduce computational cost and improve the performance of the FER module, respectively.
We believe that SFEDet may contribute to the development of efficient RGB-T object detection.

\section*{Acknowledgments}

This work was supported in part by the Guangdong Basic and Applied Basic Research Foundation (Grants No. 2025A1515010705).

\clearpage
{
    \bibliographystyle{splncs04}
    \bibliography{main}
}

\clearpage
\setcounter{page}{1}
\section*{Appendix}
\appendix


\renewcommand{\thetable}{\Alph{table}}
\renewcommand{\thefigure}{\Alph{figure}}

\section{Effect of the proposed FER }
\label{app:fer}

We conduct a comparison to evaluate the effectiveness of the proposed FER module, as shown in Table~\ref{tab:fer}. We adjust the score threshold to maintain the same recall rate as the RPNs and count the false positives in each image pair. The results demonstrate that FER more effectively suppresses false-positive proposals generated by the RPNs.
{
\setlength{\tabcolsep}{8pt}
\begin{table}[!h]
\centering
\caption{Mean average of the number of false positives in an image.}
\label{tab:fer}
\footnotesize
\begin{tabularx}{0.9\linewidth}{
>{\centering\arraybackslash}X
>{\centering\arraybackslash}X
>{\centering\arraybackslash}X
>{\centering\arraybackslash}X
>{\centering\arraybackslash}X
}
\toprule
 & & \multicolumn{3}{c}{Number of false positives}  \\ \cmidrule(lr){3-5}
\multirow{-2}{*}[2pt]{Datasets} &  \multirow{-2}{*}[2pt]{\makecell{Recall \\ Required}} & Dual RPNs   & Base  & FER   \\ \midrule
M3FD    & 0.96  & 18.0    & 26.5  & 2.1   \\
FLIR    & 0.95  & 65.0    & 38.0  & 33.5    \\
\bottomrule
\end{tabularx}
\end{table}
}

\section{Additional Details of Implementation}
\label{app:implem}

In this section, we detail the implementation and training phase. In the construction of the proposed SFEDet, we try to balance the computational cost and the detection performance. Each component in the SFEDet and the FER module is carefully designed. The parameters of each component are detailed in Table~\ref{tab:parameter}.

{
\setlength{\tabcolsep}{12pt}
\begin{table}[!ht]
\centering
\caption{Statistics for the number of parameters of each module.}
\label{tab:parameter}
\footnotesize
\begin{tabularx}{0.7\linewidth}{
>{\raggedright\arraybackslash}X
l
}
\toprule
Module & Parameters \\ \midrule
RPN-RGB (YOLOv8-Small) & 8.4M      \\
RPN-TIR (YOLOv8-Small) &  8.4M     \\
DEGF    & 0.95M          \\
Auxiliary Enhancement-1 & 1.11M         \\
FER-Head-1  & 2.27M          \\
Auxiliary Enhancement-2 & 1.11M          \\
FER-Head-2  & 2.27M          \\  \midrule
Total   & 24.5M     \\
\bottomrule
\end{tabularx}
\end{table}
}

Two independent RPNs are two standard YOLOv8-Small~\cite{yolov8} detectors, with multi-class classification retained to improve feature quality. The DEGF module, as mentioned in the Method, is a high-efficiency, lightweight fusion module. Two detection heads in the FER module are the heaviest modules in our framework. The heads share the same setting as those in YOLOv8, which adopt decoupled heads. Each head contains two independent prediction branches, \ie~the classification branch and regression branch. Though the number of channels in our SFEDet is set to 128 in most modules, the dimensions of the channels in the two heads are set to 512 for better performance. Though the prediction head is heavyweight, the total efficiency is guaranteed because of the sparse fusion.

\begin{figure*}[!t]
    \centering
    \begin{subfigure}[b]{0.31\textwidth}
        \includegraphics[width=\linewidth]{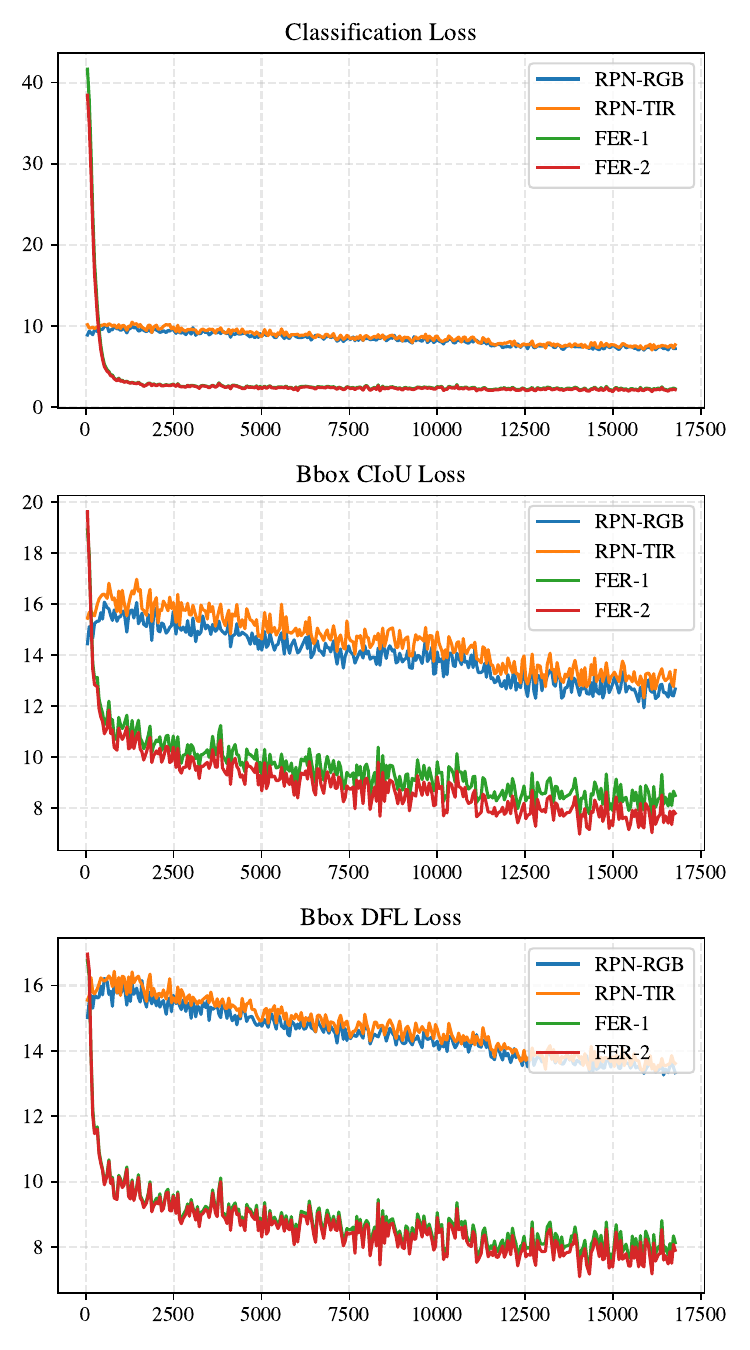} \vspace{-15pt}
        \caption{M3FD}
    \end{subfigure}
    \begin{subfigure}[b]{0.31\textwidth}
        \includegraphics[width=\linewidth]{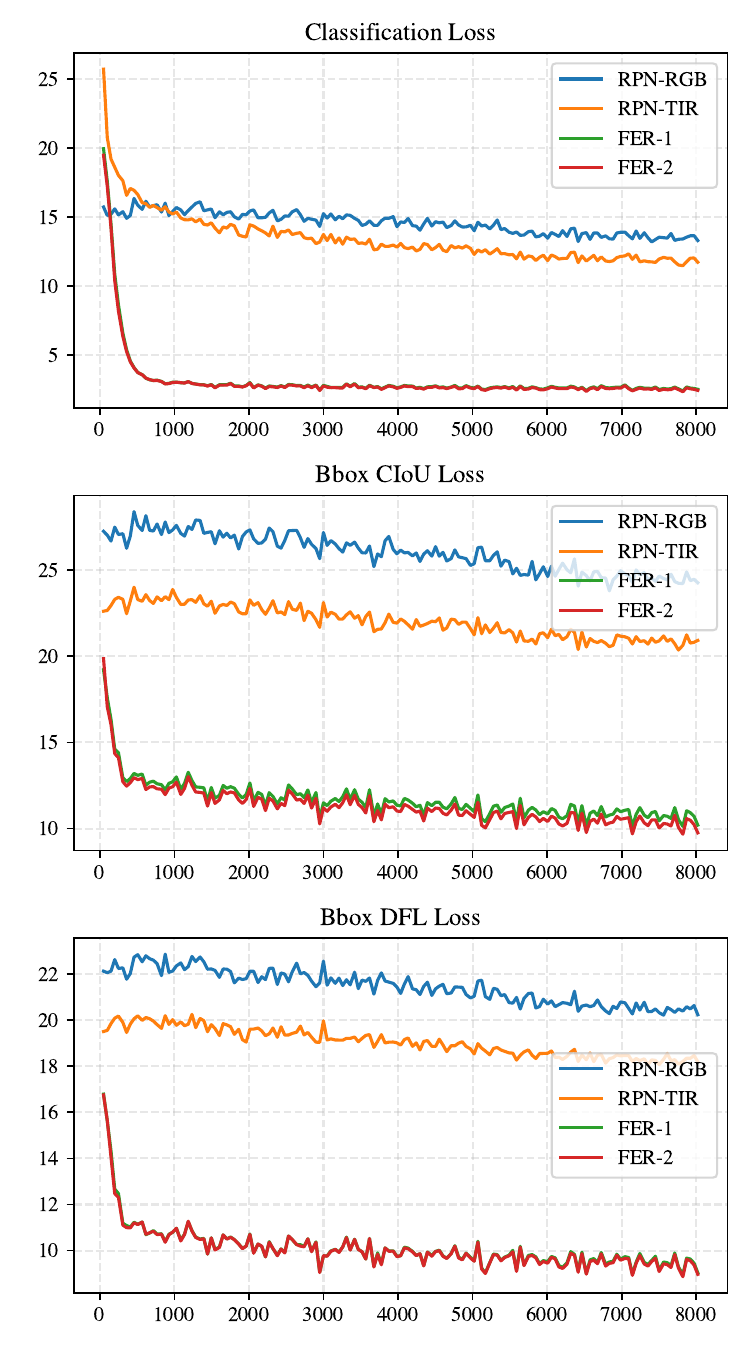} \vspace{-15pt}
        \caption{FLIR}
    \end{subfigure}
    \begin{subfigure}[b]{0.31\textwidth}
        \includegraphics[width=\linewidth]{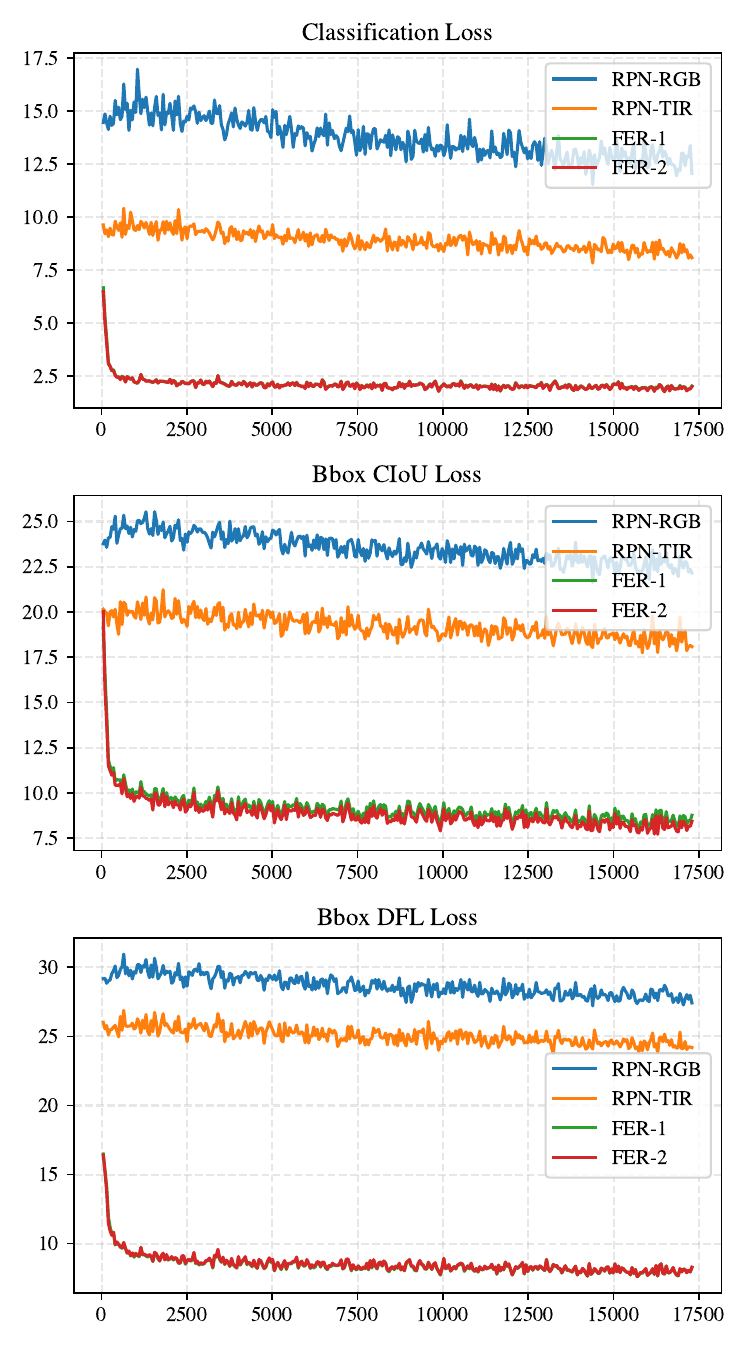} \vspace{-15pt}
        \caption{LLVIP}
    \end{subfigure} 
    \caption{Training loss curves of SFEDet in different datasets. All metrics for different detection heads are calculated under the same scale and standard. The different losses during training show the capability of different heads, where FER heads have a much stronger capacity.}
    \label{fig:loss_curves}
\end{figure*}

We adopt several techniques in training. First, we deploy the Exp-Momentum Exponential Moving Average (ExpMomentumEMA)~\cite{yolox} to achieve a better convergence in thermal-related datasets, where the momentum is set to $1e-4$. At the beginning of training, we deploy a linear warm-up scheduler for the learning rate for 500 training iterations. Then, we decay the learning rate by a 0.1 factor after different training epochs for different datasets for better convergence.
Borrowing from literature~\cite{Tian_Yang_Zhu_Wang_He_2025}, we randomly add noise to one of the modalities during RGB and TIR. The type of noise is randomly selected during \textit{Gaussian}, \textit{Contrast}, and \textit{Blur}, following a uniform distribution. These data augmentations enhance the diversity of samples, thus improving the robustness of features after fusion and overcoming the modality imbalance issue~\cite{Tian_Yang_Zhu_Wang_He_2025}.

\vspace{1mm} \noindent
\textbf{Training loss analysis.} 
The loss curves of SFEDet on different datasets are shown in Figure~\ref{fig:loss_curves}. Across all datasets, the losses of the FER heads exhibit a sharp decrease at the beginning, as the FER module is trained from scratch. During the later training stages, the deeper heads tend to converge to lower loss values under the same evaluation criteria, which suggests an improvement in the representational capacity of the detector.
Furthermore, the differences between the losses of the two RPNs reflect the distinct contributions of the RGB and TIR modalities. In the M3FD~\cite{m3fd} dataset, the RGB and TIR modalities tend to contribute comparably to the overall detection performance, owing to their complementary sensitivities to different object categories, such as traffic lights (in RGB) and persons (in TIR). In contrast, for the FLIR~\cite{zhang2020multispectral_flir} and LLVIP~\cite{llvip} datasets, where the primary target objects (i.e., persons and cars) are generally more salient in TIR images, the TIR modality plays a more dominant role.
The comparison between RPNs and the final heads demonstrates the effectiveness of the proposed cross-modality sparse fusion and examination.

\section{More Qualitative Detection Results}

\begin{figure}[!h]
\centering
\includegraphics[width=\linewidth]{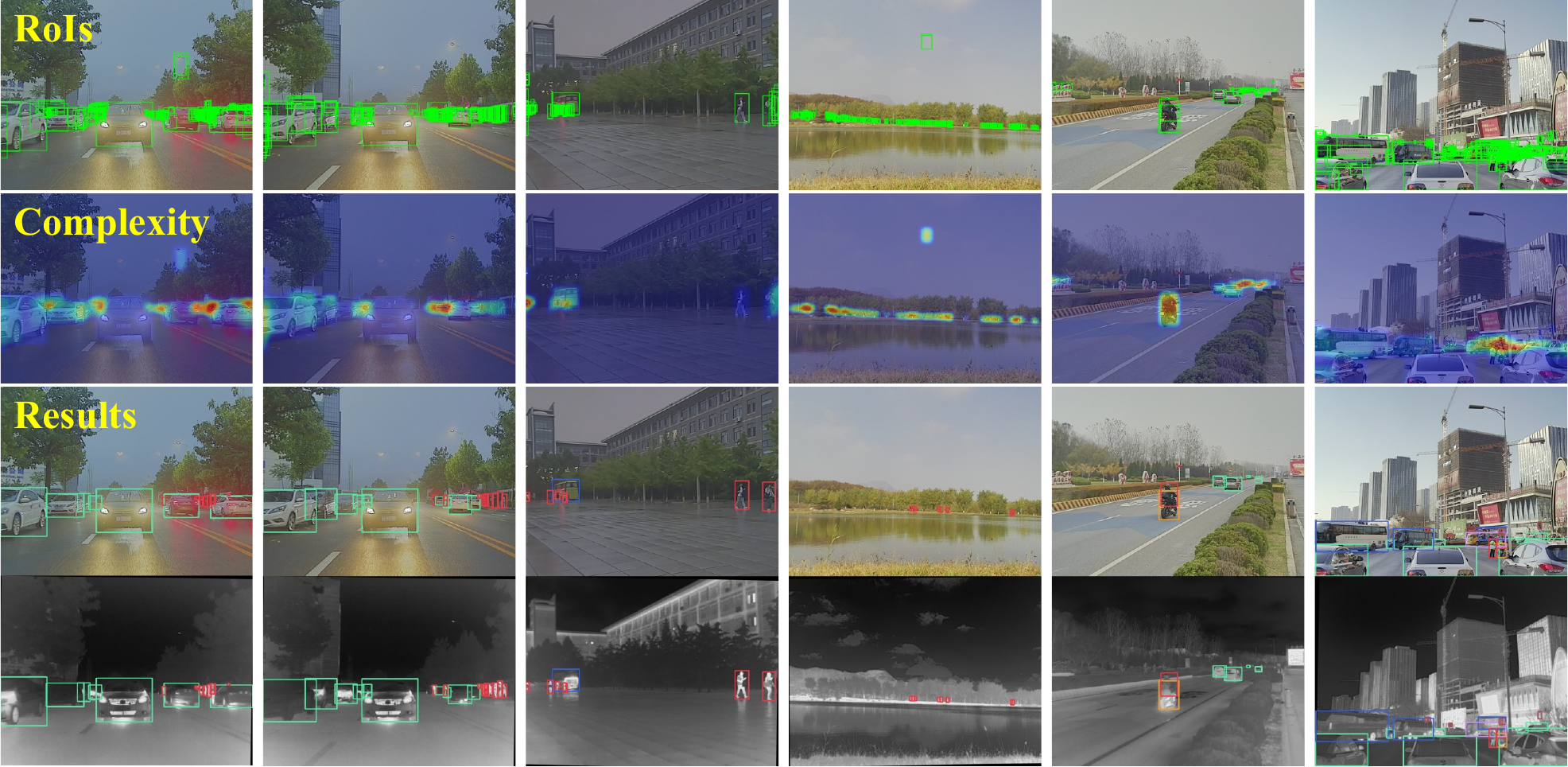}
\caption{Qualitative detection results on M3FD dataset. Different rows respectively show the corresponding RoIs, complexity, and results. Different categories are marked in different colors.}
\label{fig:rois_m3fd}
\end{figure}

\begin{figure}[!th]
\centering
\includegraphics[width=\linewidth]{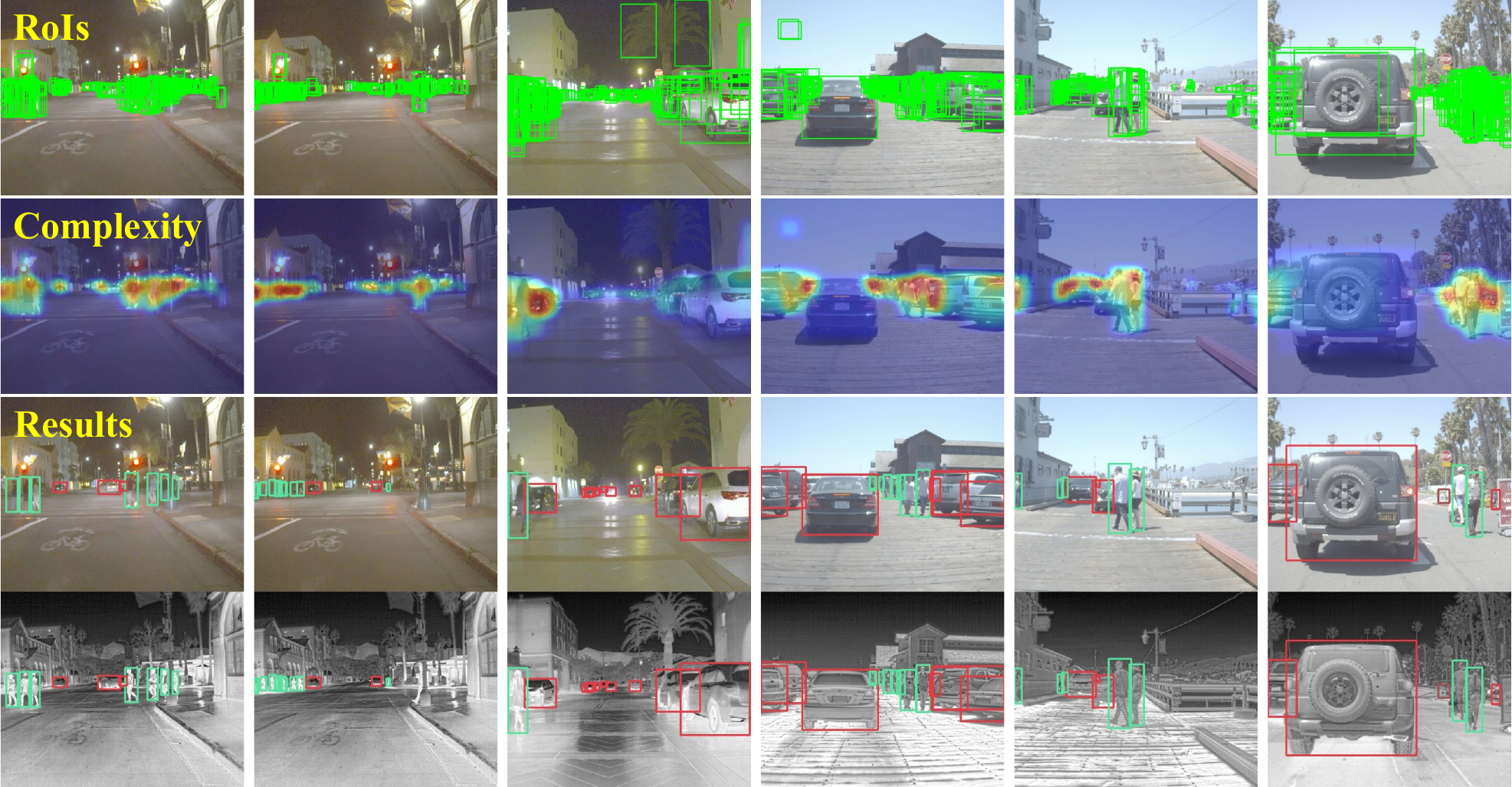}
\caption{Qualitative detection results on FLIR dataset. Due to the poor imaging quality in RGB images of the FLIR dataset, the proposals obtained from the RGB are a bit more than those of other datasets.}
\label{fig:rois_flir}
\end{figure}

\begin{figure}[!th]
\centering
\includegraphics[width=\linewidth]{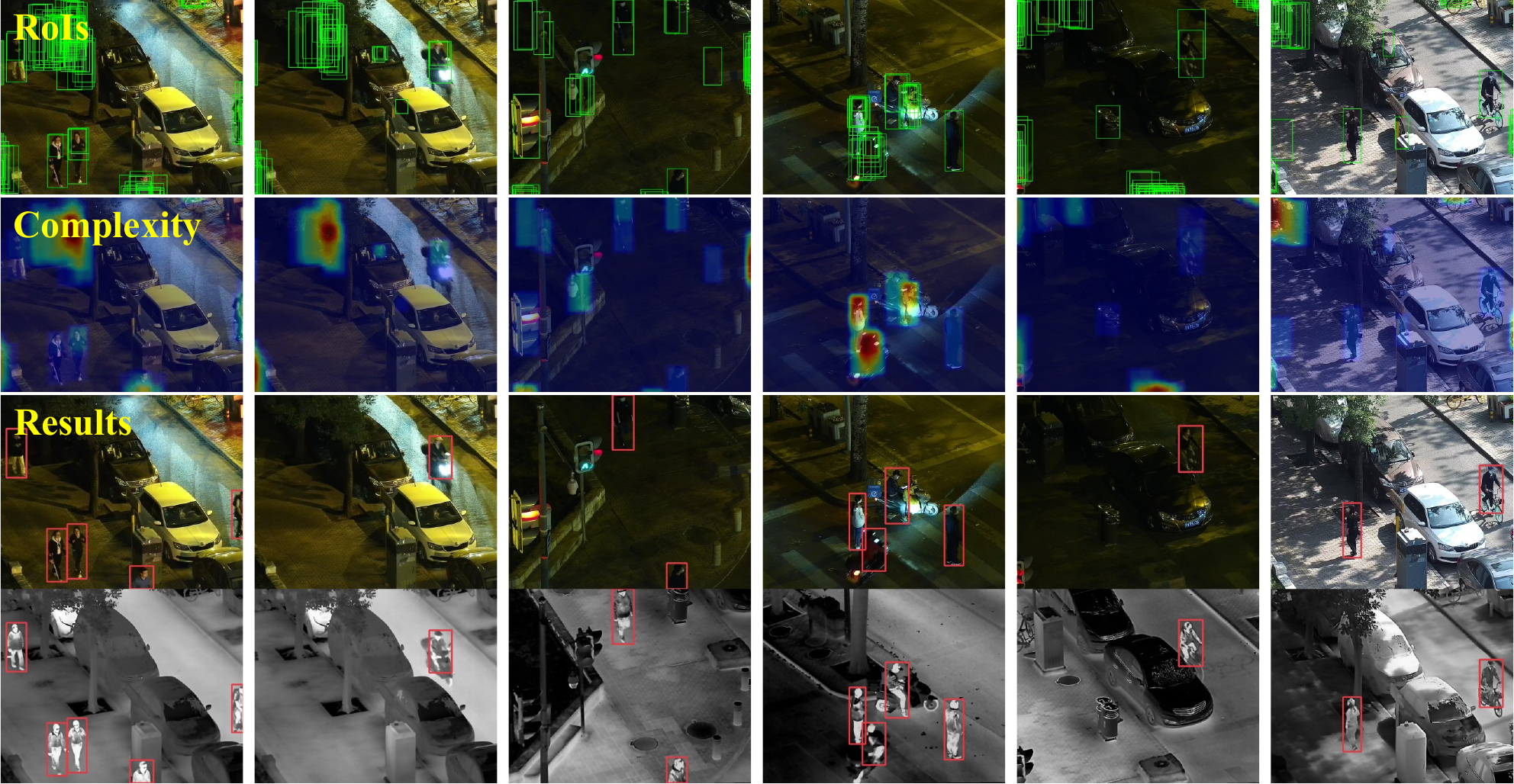}
\caption{Qualitative detection results on LLVIP dataset. According to the above results, most of the background areas are pre-filtered, avoiding the fusion and heavyweight inference module. In the second stage, the RoIs are carefully examined and refined to find out the sparse bounding boxes.}
\label{fig:rois_llvip}
\end{figure}

We present more qualitative detection results of our SFEDet. The results from different datasets are respectively shown in Figure~\ref{fig:rois_m3fd}, Figure~\ref{fig:rois_flir} and Figure~\ref{fig:rois_llvip}. The first row of each figure is the RoIs obtained from RPNs, where the preliminary detections are processed by the NMS to get the sparse proposals. The second row shows the distribution of computational complexity. Considering that some regions are repetitively computed, these regions suffer a heavier complexity, highlighted with the hot color. The third row shows the final detection bounding boxes, where the boxes of different colors mean different categories. 

The sparsity of RoIs demonstrates the effectiveness of our SFEDet. 
Due to the poor imaging quality in RGB images of the FLIR dataset, the proposals obtained from the RGB images are slightly more than those from other datasets. Meanwhile, in the LLVIP dataset, due to the lower challenging conditions, most of the background areas are pre-filtered, avoiding the fusion and heavyweight inference module. In the second stage, the RoIs are carefully examined and refined to find the sparse bounding boxes. The difference between RoIs and the final results shows the achievement of our proposed fusion-driven examination and refinement module, which excludes the false positives in proposals and modifies the inaccurate bounding boxes.


\end{document}